\documentclass{article}

\usepackage{microtype}
\usepackage{graphicx}

\usepackage{caption}

\usepackage{booktabs} % for professional tables

\usepackage[colorlinks,citecolor=green]{hyperref}

\usepackage[accepted]{icml2024}
\usepackage{amsmath}
\usepackage{amssymb}
\usepackage{mathtools}
\usepackage{amsthm}

\usepackage[capitalize,noabbrev]{cleveref}

\usepackage{listings}
\usepackage{pythonhighlight}
\usepackage{fancyvrb}
\RecustomVerbatimCommand{\VerbatimInput}{VerbatimInput}{fontsize=\footnotesize,
 frame=single,  
 framesep=0.5em, % separation between frame and text
 labelposition=topline,
}

\usepackage{tcolorbox}
\tcbuselibrary{listings}
\usepackage{bbding}
\usepackage{pifont}
\definecolor{green}{HTML}{009B55}
\usepackage{xspace}
\usepackage{macros}
\usepackage{wrapfig}
\usepackage{lipsum}
\usepackage{url}            % simple URL typesetting
\usepackage{amsfonts}       % blackboard math symbols
\usepackage{nicefrac}       % compact symbols for 1/2, etc.
\usepackage{xcolor}         % colors
\usepackage[colorlinks,citecolor=green]{hyperref}
\usepackage{multicol}
\usepackage{multirow}

\usepackage{listings}
\usepackage{pythonhighlight}
\usepackage{subfigure}
\usepackage{scalerel}
\usepackage{Definitions}
\usepackage{algorithm}
\usepackage{algorithmic}
\usepackage{placeins}

\definecolor{nblue}{cmyk}{0.95,0.0,0.2,0.2}

\usepackage{xcolor, colortbl}
\usepackage{makecell}
\usepackage{tablefootnote}
\newcommand{\method}{\textsc{BBox-Adapter}\xspace}

\theoremstyle{plain}

\theoremstyle{definition}

\theoremstyle{remark}

\icmltitlerunning{Lightweight Adapting for Black-Box Large Language Models}

\begin{document}

\twocolumn[
\icmltitle{ \method: Lightweight Adapting for Black-Box Large Language Models}



\icmlsetsymbol{equal}{*}

\begin{icmlauthorlist}
\icmlauthor{Haotian Sun\textsuperscript{*}}{gt}
\icmlauthor{Yuchen Zhuang\textsuperscript{*}}{gt}
\icmlauthor{Wei Wei}{ac}
\icmlauthor{Chao Zhang}{gt}
\icmlauthor{Bo Dai}{gt}
\end{icmlauthorlist}

\icmlaffiliation{gt}{Georgia Tech}
\icmlaffiliation{ac}{Accenture}

\icmlcorrespondingauthor{Haotian Sun}{haotian.sun@gatech.edu}
\icmlcorrespondingauthor{Bo Dai}{bodai@cc.gatech.edu}

\icmlkeywords{Machine Learning, ICML}

\vskip 0.3in
]



\printAffiliationsAndNotice{\icmlEqualContribution} % otherwise use the standard text.

\newcommand{\version}{Camera_ready/}

\begin{abstract}
  Adapting state-of-the-art Large Language Models (LLMs) like GPT-4 and Gemini for specific tasks is challenging.
  Due to the opacity in their parameters, embeddings, and even output probabilities,
  existing fine-tuning adaptation methods are inapplicable.
  Consequently, adapting these black-box LLMs is only possible through their API services, raising concerns about transparency, privacy, and cost.
  To address these challenges, we introduce \method, a novel lightweight adapter for black-box LLMs. \method distinguishes target and source domain data by treating target data as positive and source data as negative.
  It employs a ranking-based Noise Contrastive Estimation (NCE) loss to promote the likelihood of target domain data while penalizing that of the source domain.
  Furthermore, it features an online adaptation mechanism, which incorporates real-time positive data sampling from ground-truth, human, or AI feedback, coupled with negative data from previous adaptations.
  Extensive experiments demonstrate \method's effectiveness and cost efficiency.
  It improves model performance by up to $6.77\%$ across diverse tasks and domains, while reducing training and inference costs by $31.30$x and $1.84$x, respectively.
\end{abstract}

\begin{table*}[t]
% \floatconts
\centering
  \caption{
  Comparison of existing LLM adaptation methods based on five aspects: (1) Model parameters accessibility, (2) Access to high-dimensional representations of input sequences or output generations, (3) Token probability availability, (4) Retrieval corpus necessity, and (5) Utilization of a smaller adapter model. 
}
  \label{tab:baseline_attr}
  \fontsize{8}{10}\selectfont \setlength{\tabcolsep}{0.4em}
  \begin{tabular}{l|ccc|cc}
  \toprule
  \bfseries Methods & \bfseries \makecell{w/o Model\\ Parameters} & \bfseries \makecell{w/o High-Dimensional\\ Representation} & \bfseries \makecell{w/o Token\\ Probabilities} & \bfseries \makecell{w/o Retrieval\\ Corpus} & \bfseries \makecell{w/ Smaller\\ Adapter}\\
  \midrule
  \multicolumn{6}{l}{\quad \emph{White-Box LLM Fine-Tuning}}  \\\midrule 
  Fine-Tuning~\cite{devlin-etal-2019-bert} & \color{red}{\ding{55}} & \color{red}{\ding{55}} & \color{red}{\ding{55}} & \color{green}{\ding{51}} & \color{red}{\ding{55}}\\
  Instruction-Tuning~\cite{wei2021finetuned} & \color{red}{\ding{55}} & \color{red}{\ding{55}} & \color{red}{\ding{55}} & \color{green}{\ding{51}} & \color{red}{\ding{55}}\\
  Continual Pre-Training~\cite{gururangan-etal-2020-dont} & \color{red}{\ding{55}} & \color{red}{\ding{55}} & \color{red}{\ding{55}} & \color{green}{\ding{51}} & \color{red}{\ding{55}}\\
  % \midrule
  % \multicolumn{6}{l}{\quad \emph{Parameter-Efficient Fine-Tuning}}  \\\midrule 
  Adapter~\cite{houlsby2019parameter} & \color{red}{\ding{55}} & \color{red}{\ding{55}} & \color{red}{\ding{55}} & \color{green}{\ding{51}} & \color{green}{\ding{51}}\\
  Prefix-Tuning~\cite{liu-etal-2022-p}& \color{red}{\ding{55}} & \color{red}{\ding{55}} & \color{red}{\ding{55}} & \color{green}{\ding{51}} & \color{green}{\ding{51}}\\
  LoRA~\cite{hu2021lora}& \color{red}{\ding{55}} & \color{red}{\ding{55}} & \color{red}{\ding{55}} & \color{green}{\ding{51}} & \color{green}{\ding{51}}\\\midrule
  \multicolumn{6}{l}{\quad \emph{Grey-Box LLM Adaptation}}  \\\midrule 
  LMaaS~\cite{sun2022black} & \color{green}{\ding{51}} & \color{red}{\ding{55}} & \color{red}{\ding{55}} & \color{green}{\ding{51}} & \color{green}{\ding{51}}\\
  kNN-Adapter~\cite{huang2023k} & \color{green}{\ding{51}} & \color{green}{\ding{51}} & \color{red}{\ding{55}} & \color{red}{\ding{55}} & \color{green}{\ding{51}}\\
  CombLM~\cite{ormazabal-etal-2023-comblm} & \color{green}{\ding{51}} & \color{green}{\ding{51}} & \color{red}{\ding{55}} & \color{green}{\ding{51}} & \color{green}{\ding{51}}\\
  IPA~\cite{lu2023ipa} & \color{green}{\ding{51}} & \color{green}{\ding{51}} & \color{red}{\ding{55}} & \color{green}{\ding{51}} & \color{green}{\ding{51}}\\
  Proxy-Tuning~\cite{liu2024tuning} & \color{green}{\ding{51}} & \color{green}{\ding{51}} & \color{red}{\ding{55}} & \color{green}{\ding{51}} & \color{green}{\ding{51}}\\\midrule
  \multicolumn{6}{l}{\quad \emph{Black-Box LLM Adaptation}}  \\\midrule
  \rowcolor{teal!10} \textbf{\method (Ours) } & \color{green}{\ding{51}} & \color{green}{\ding{51}} & \color{green}{\ding{51}} & \color{green}{\ding{51}} & \color{green}{\ding{51}} \\
  \bottomrule
  \end{tabular}
  \vspace{-2ex}
\end{table*}

\section{Introduction}\label{sec:intro}

\begin{figure}[t]
  \centering
  \includegraphics[width=0.95\linewidth]{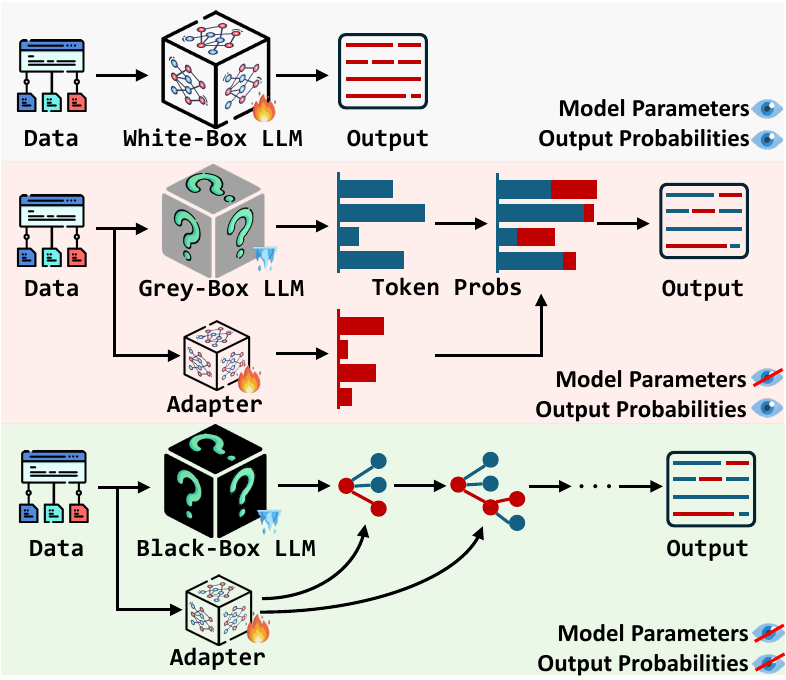}
  \caption{Illustration of white-box, grey-box, and black-box LLM adaptation. White-box has complete access to both model parameters and output probabilities, grey-box has access only to output probabilities, and black-box lacks access to both.
    \raisebox{-0.18em}{\includegraphics[height=1em]{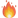}} indicates the models with trainable parameters, whereas \raisebox{-0.18em}{\includegraphics[height=1em]{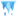}} indicates the inaccessible fixed parameters.
  }
  \vspace{-3ex}
  \label{fig:teaser}
\end{figure}

Large Language Models (LLMs) have demonstrated exceptional abilities in comprehending and generating text across a wide range of tasks~\cite{radfordimproving, radfordlanguage, brown2020language, openai2023, chowdhery2022palm}.
Despite their growing capabilities, general-purpose, pre-trained LLMs still require further customization to achieve optimal performance on specific use cases.
However, adapting \emph{black-box} LLMs like GPT-3.5~\cite{chatgpt} and Gemini~\cite{team2023gemini} presents significant challenges due to the lack of direct access to internal model parameters.

Adapting black-box LLMs can be achieved by preparing and uploading training data through fine-tuning APIs, such as the OpenAI GPT-3.5-turbo fine-tuning API~\cite{SFTgpt}. However, employing fine-tuning APIs for LLM adaptation has several critical issues:
(1) \textbf{Transparency}: Aside from a restricted set of adjustable hyperparameters (\eg, the number of tuning epochs), the fine-tuning process remains largely opaque. Crucial aspects, such as the extent of trainable layers and specific model weights, are often undisclosed, hindering optimal customization.
(2) \textbf{Privacy}: Uploading training data via APIs introduces potential risks of privacy breaches, limiting the use of LLMs in sensitive domains. For instance, electronic health records containing confidential healthcare information require stringent privacy measures.
(3) \textbf{Cost}: The cost associated with fine-tuning APIs is considerably higher compared to inference, making the adaptation expensive. The fine-tuning cost will significantly increase with hyperparameter tuning. 

The adaptation of black-box LLMs without the use of APIs remains an \emph{unresolved} challenge.
Recent studies have explored adapting LLMs without accessing model weights, by integrating outputs with tunable white-box models~\cite{sun2022black,ormazabal-etal-2023-comblm, lu2023ipa, liu2024tuning} or external data sources~\cite{huang2023k}.
However, such approaches (depicted as \emph{grey-box} adaptation in Figure~\ref{fig:teaser}) still require access to the token probabilities of the output sequences, only available in models preceding GPT-3~\cite{brown2020language} or white-box LLMs like LLaMA-2~\cite{touvron2023llama}.
Output probabilities, unfortunately, are inaccessible in recent black-box LLMs~\footnote{We explain the inaccessibility of output token probabilities in state-of-the-art black-box LLMs in Appendix~\ref{app:token}.} like GPT-3.5~\cite{chatgpt} and PaLM-2~\cite{anil2023palm2}, making these techniques inapplicable for state-of-the-art black-box LLMs.

We propose \method, a lightweight adapter that adapts black-box LLMs for specific tasks by fine-tuning a smaller language model (LM) with just 0.1B-0.3B parameters.
We formulate the black-box LLM adaptation process as a sampling problem from an energy-based model (EBM).
To effectively distinguish between source and target domain data, we design a ranking-based noise contrastive estimation (NCE) loss for adapter updates.
We combine outputs from the black-box LLM and the adapter for adaptive inference. \method employs an online adaptation framework, iteratively sampling from previous inferences and updating the adapter.
Notably, the adapter facilitates self-improvement through AI feedback during training, reducing the reliance on ground-truth training data as positive samples in the online adaptation process.

Extensive experiments across three diverse datasets demonstrate the effectiveness of \method in adapting black-box LLMs to downstream tasks, achieving performance gains of up to $6.77\%$, while significantly reducing training and inference costs of fine-tuning methods. Moreover, \method accomplishes black-box LLM adaptation without requiring access to model parameters or output probabilities, enabling transparent, privacy-conscious, and cost-effective customization of cutting-edge LLMs.
We summarize the main contributions as follows:

$\bullet$ We first categorize the adaptation methods systematically based on the accessible information for the algorithms. 

$\bullet$ We introduce \method, a novel energy-based adapter that fine-tunes a smaller LM to facilitate black-box LLM adaptation without fine-tuning APIs. To the best of our knowledge, \method is the first black-box adapter to enable state-of-the-art LLM (\eg, GPT-3.5) adaptation without model weights or output probabilities.

$\bullet$ \method is lightweight, using a small model with just 0.1B-0.3B parameters as the adapter. It surpasses supervised fine-tuning (SFT) by 31.30 times during training and 1.84 times during inference in terms of cost.

$\bullet$ \method is also applicable without ground-truth data for the task. Its online adaptation framework can use negative samples from previous model inferences and positive samples from various sources, including AI feedback. This allows \method to remain effective even when ground-truth data is limited or unavailable.

$\bullet$ \method offers a generalizable and flexible solution for LLM adaptation. It can be applied to a wide range of tasks, domains, and models of varying sizes. Once the adapter is tuned for a specific task or domain, it can be directly applied to other black-box LLMs in a plug-and-play manner, eliminating the need for further retraining.

\section{Categorization of LLM Adaptation}

Based on the accessibility to internal model parameters and output probabilities, we categorize LLM adaptation methods into three main groups (Table~\ref{tab:baseline_attr}): \emph{white-box} fine-tuning (full access), \emph{grey-box} adaptation (access to output probabilities only), and \emph{black-box} adaptation (no access). 

\textbf{White-Box LLM Fine-Tuning.}
To fully leverage the capabilities of LLMs in language comprehension and enhance their performance, many users still need to customize them for specific tasks and domains~\cite{chung2022scaling}.
A straightforward approach to achieve this involves fine-tuning~\cite{wei2021finetuned, wang-etal-2022-super} or continuous pre-training~\cite{ke2022continual, gupta2023continual} 
the LM on domain-specific data. However, these methods require extensive computational resources and memory, which becomes increasingly challenging as model sizes grow exponentially.
To mitigate the computational and memory burdens for LLM fine-tuning, Parameter-Efficient Fine-Tuning (PEFT) methods~\cite{hu2021lora, houlsby2019parameter, he2021towards, li2021prefix} have been proposed that focus on training only a small subset of parameters rather than the entire model. 
Examples of such techniques include adapters~\cite{houlsby2019parameter}, prefix tuning~\cite{liu-etal-2022-p,li2021prefix}, and low-rank adaptation~\cite{hu2021lora}.
Unfortunately, these techniques require direct access to the internal parameters of the original model and complete backward passes, making them incompatible with black-box models.

\textbf{Grey-Box LLM Adaptation.}
For grey-box LLM adaptation, existing approaches make different assumptions about the transparency of the LLM.
One line of research assumes that only the gradient information is unavailable, while the high-dimensional input and output sequences are accessible.
For example, LMaaS~\cite{sun2022black} trains a small, derivative-free optimizer for discrete prompt tuning to enhance the probabilities of ground-truth tokens from the target domain.
Another line of research assumes that only output token probabilities from black-box LLMs are available.
kNN-Adapter~\cite{huang2023k} augments a black-box LLM with k-nearest neighbor retrieval from an external, domain-specific datastore. It adaptively interpolates LM outputs with retrieval results from the target domain.
CombLM~\cite{ormazabal-etal-2023-comblm} employs fine-tuning on a smaller white-box model to align the output token probabilities of a black-box LLM with the target distribution.
Similarly, proxy-tuning~\cite{liu2024tuning} fine-tunes a smaller LM as an `expert' while its untuned version serves as an `anti-expert'. The method involves adjusting the black-box LLM outputs by adding the logit offsets from their token-level predictions for adaptation.
CaMeLS~\cite{hu2023meta} meta-trains a compact, autoregressive model to dynamically adjust the language modeling loss for each token during online fine-tuning.
However, these methods are inapplicable to the latest state-of-the-art black-box LLMs, such as GPT-4~\cite{openai2023} and PaLM2~\cite{anil2023palm2}, due to the inaccessibility of token probabilities.

\textbf{Black-Box LLM Adaptation.}
Due to the black-box nature, users are unable to access 
(1) internal model parameters, (2) high-dimensional representations of input sequences or output generations, and (3) output token probabilities for their specific use cases in black-box adaptation.
Notably, existing methods, except ours, fail to support \emph{black-box} LLM adaptations, where neither model parameters nor output probabilities can be accessed in most recent LLMs like GPT-3.5~\cite{chatgpt} and Gemini~\cite{team2023gemini}.

\section{Method}

\begin{figure*}[t]
  \centering
  \includegraphics[width=0.99\linewidth]{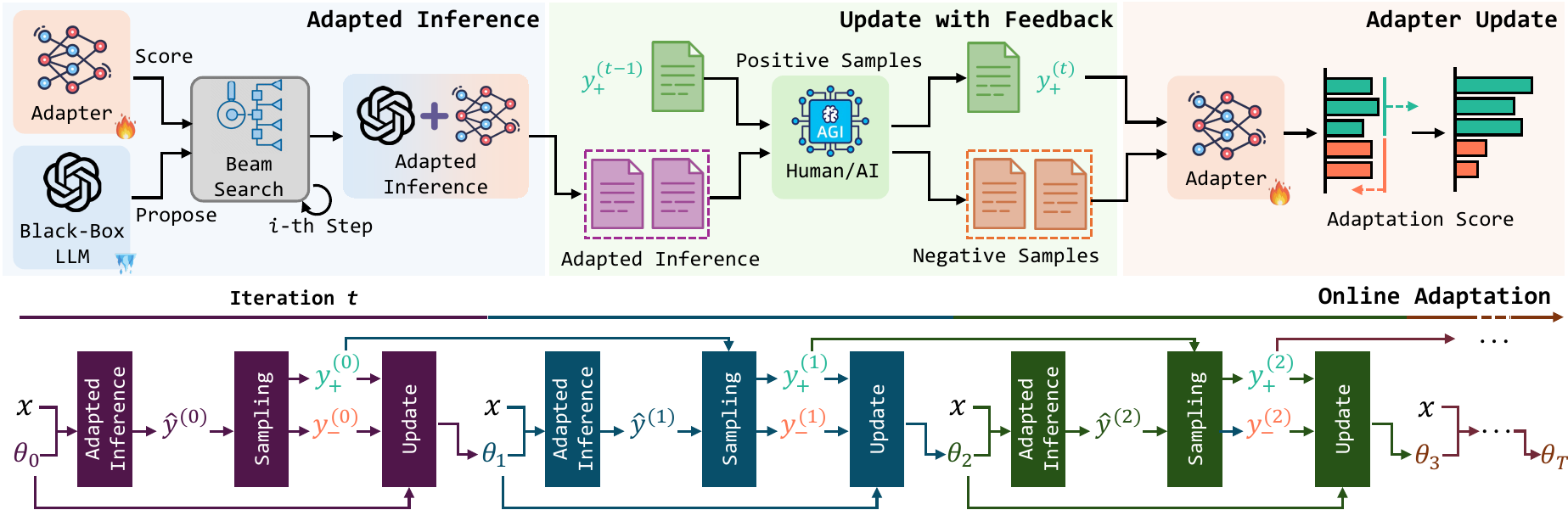}
  \caption{Overview of \method for black-box LLM adaptation from the source to the target domain. 
  \method adopts an online adaptation framework, iteratively sampling from previous inferences and updating the adapter.
  }
  \label{fig:overview}
\end{figure*}

In this section, we present \method, a lightweight method for adapting black-box LLMs to specific tasks  (Figure~\ref{fig:overview}).
We first frame the black-box LLM adaptation process as a sampling problem from an EBM (Section~\ref{subsec:energy}).
Following this EBM perspective, we derive a ranking-based NCE loss for adapter updates (Section~\ref{subsec:ada-update}), enabling the distinction between source and target domain data.
We then describe the process of combining outputs from the black-box LLM and the adapter for adapted inference (Section~\ref{subsec:inference}).
To model the real distributions of both source and target domains, we introduce \method as an online adaptation framework that iteratively samples from the previously adapted inferences and updates the adapters accordingly (Section~\ref{subsec:method-oa}).

\subsection{Black-Box LLM Adaptation as EBM}\label{subsec:energy}

To effectively adapt a black-box LLM, our objective is to calibrate its output generation from the original source domain to align with a specific target domain. 
This process involves conceptualizing the source and target domains as distributions within a joint space, \(\mathcal{Y} \sim \mathcal{Y}^S \times \mathcal{Y}^T\), where \(\mathcal{Y}^S\) and \(\mathcal{Y}^T\) represent the text generations of the source and target domains, respectively.
Specifically, given a target domain dataset \(\mathcal{D} = \{(\mathbf{x}_i, \mathbf{y}^t_i)\}_{i=1}^N\), our goal is to steer the output of the black-box LLM \(\hat{\mathbf{y}}_i\) towards a transition from the source domain output \(\hat{\mathbf{y}}^s_i \in \mathcal{Y}^S\) to the target domain's ground-truth response \(\mathbf{y}_i^t \in \mathcal{Y}^T\) for each input sequence \(\mathbf{x}_i\). 
This transition is crucial to ensuring that the model's outputs become more tailored to the desired target domain.

We frame black-box LLMs adaptation as a problem of sampling from a specialized energy-based sequence model \(p_\theta\). This model defines a globally normalized probability distribution that satisfies the desired constraints we aim to integrate during the adaptation process. Consequently, we can parameterize the distribution of the adaptation as follows:
\begin{equation}
    \begin{aligned}
        p_\theta(\mathbf{y}|\mathbf{x})=p_{\text{LLM}}(\mathbf{y}|\mathbf{x})\frac{\exp(g_\theta(\mathbf{x},\mathbf{y}))}{Z_\theta(\mathbf{x})},
    \end{aligned}\label{eq:adaptation}
\end{equation}
where \(Z_\theta(\mathbf{x})=\int p_{\text {LLM}}(\mathbf{y} | \mathbf{x}) \exp(g_\theta(\mathbf{x},\mathbf{y}))d\mathbf{y}\) is the normalizing factor known as the partition function, \(p_\theta\) denotes the adapted model, \(p_{\text{LLM}}\) remains fixed as the black-box model, and \(g_\theta\) represents the adapter.
The goal of training is to learn the adapter's parameters such that the joint model distribution approaches the data distribution. For notation clarity, we will omit the conditioning variables in the subsequent discussion. Thus, the equation above can be rewritten as \(p_\theta(\mathbf{x})=p_{\text{LLM}}(\mathbf{x})\frac{\exp(g_\theta(\mathbf{x}))}{Z(\theta)}\).

\subsection{Adapter Update}\label{subsec:ada-update}

As $Z(\theta)$ is intractable, the maximum likelihood estimation (MLE) of $p_\theta(\mathbf{x})$ requires either sampling from the model distributions or approximation operations, which are computationally intensive and often imprecise.
To address this, we employ NCE \citep{gutmann2010noise,ma2018noise,oord2018representation,deng2020residual} as an efficient estimator for $g_\theta\rbr{\xb}$.
Our approach extends beyond the conventional NCE, which only categorizes samples as either `real' or `noise'. Instead, we employ a ranking-based NCE loss that prioritizes ranking true data samples higher than noise~\citep{ma2018noise}. 
We denote the posterior $q(k|\{\mathbf{x}_k\}_{k=1}^K)$ to be $q(x_k\text{ is positive} | \{x_k\}_{k=1}^K)$. Specifically, this denotes the probability that the $k$-th sample is drawn from the ground-truth dataset. Here $[x_k\text{ is positive}]$ is the indicator of $x_k$ being the positive sample. Similarly, we apply the simplified notation on $p_\theta(k|\{\mathbf{x}_k\}_{k=1}^K)$.
Assuming the auxiliary label differentiates between a positive sample from data and a negative one from the LLM, we consider the samples $\{\mathbf{x}_k\}_{k=1}^K$ to estimate the posterior of the label distribution:
\begin{equation*}
\begin{small}
    \begin{aligned}
        q(k|\{\mathbf{x}_k\}_{k=1}^K)=\frac{p_{\text{data}}(\mathbf{x}_k)\prod_{i\neq k}p_{\text{LLM}}(\mathbf{x}_i)}{\sum_kp_{\text{data}}(\mathbf{x}_k)\prod_{i\neq k}p_{\text{LLM}}(\mathbf{x}_i)}=\frac{\frac{p_{\text{data}}(\mathbf{x}_k)}{p_{\text{LLM}}(\mathbf{x}_k)}}{\sum_k\frac{p_{\text{data}}(\mathbf{x}_k)}{p_{\text{LLM}}(\mathbf{x}_k)}}.
    \end{aligned}
    \end{small}
\end{equation*}
We can parameterize $p_\theta(k|\{\mathbf{x}_k\}_{k=1}^K)$ as:
\begin{equation*}
    \begin{aligned}
        p_\theta(k|\{\mathbf{x}_k\}_{k=1}^K)=\frac{\exp(g_\theta(\mathbf{x}_k))}{\sum_k\exp(g_\theta(\mathbf{x}_k))}.
    \end{aligned}
\end{equation*}
By minimizing the KL-divergence between $p_\theta(k|\{\mathbf{x}_k\}_{k=1}^K)$ and $q(k|\{\mathbf{x}_k\}_{k=1}^K)$, we can frame the problem as:
% \vspace{-2ex}
\begin{equation}
    \begin{aligned}
        \min_\theta\ell(\theta)=\max_\theta\mathbb{E}_{p_{\text{data}}(\mathbf{x})}[g_\theta(\mathbf{x})-\log\sum_k\exp(g_\theta(\mathbf{x}_k))].
    \end{aligned}\label{eq:kl1}
\end{equation}
We then have the optimal $\theta$ satisfies:
\begin{equation*}
    \begin{aligned}
p_\theta(k|\{\mathbf{x}_k\}_{k=1}^K) = q(k|\{\mathbf{x}_k\}_{k=1}^K),
    \end{aligned}
\end{equation*}
which implies, 
\begin{equation*}
    \begin{aligned}
        p_\theta(\mathbf{x}):=p_{\text{LLM}}(\mathbf{x}){\exp(g_\theta(\mathbf{x}))}=p_{\text{data}}(\mathbf{x}).
    \end{aligned}
\end{equation*}

Arbitrary energy models based on outputs, such as $g_\theta$, may experience sharp gradients, leading to instability during training.
To address this, we incorporate spectral normalization \citep{du2019implicit_ebm} to Eq.\eqref{eq:kl1}.
Consequently, we can derive the gradient of the loss function as follows:
\begin{equation*}
    \begin{aligned}
        \nabla_\theta\ell(\theta)=\nabla_\theta&\{-\mathbb{E}_{p_{\text{data}}}[g_\theta(\mathbf{x})]+\mathbb{E}_{p_\theta}[g_\theta(\mathbf{x})]+\alpha\mathbb{E}[g_\theta(\mathbf{x})^2]\}.
    \end{aligned}\label{eq:loss}
\end{equation*}
Considering the complete format of Eq.(\ref{eq:adaptation}), we can rewrite the gradient as:
\begin{equation}
    \begin{aligned}
        \nabla_\theta\ell(\theta)=&\nabla_\theta \{-\mathbb{E}_{\mathbf{y}_+\sim p_{\text{data}}(\mathbf{y}|\mathbf{x})}[g_\theta(\mathbf{x},\mathbf{y}_+)]+\alpha\mathbb{E}[g_\theta(\mathbf{x,\mathbf{y}_+})^2]\\
        &+\mathbb{E}_{\mathbf{y}_-\sim p_\theta(\mathbf{y}|\mathbf{x})}[g_\theta(\mathbf{x},\mathbf{y}_-)]+\alpha\mathbb{E}[g_\theta(\mathbf{x,\mathbf{y}_-})^2]\}.
    \end{aligned}\label{eq:final_loss}
\end{equation}

\subsection{Adapted Inference}\label{subsec:inference}
During model inference, we conceptualize the black-box LLM as a proposal generator, while the adapter serves as an evaluator. 
This framework allows us to decompose complicated tasks, such as multi-step reasoning and paragraph generation, into a more manageable sentence-level beam search process.
The complete solution $\mathbf{y}$ is sequentially generated at the sentence level over several time steps, represented as $\mathbf{y}=[\mathbf{s}^1,\mathbf{s}^2,\cdots,\mathbf{s}^L]=\mathbf{s}^{1:L}$, where $\mathbf{s}^l$ denotes the $l$-th sentence in the generation sequence.
We can then factorize the adapted inference process $p_\theta(\mathbf{y}|\mathbf{x})$ in an autoregressive manner:
\begin{equation*}
    \begin{aligned}
        p_{\theta}(\mathbf{y}|\mathbf{x})&=p_{\theta}(\mathbf{s}^{1:L}|\mathbf{x})=p_{\text{LLM}}(\mathbf{s}^{1:L}|\mathbf{x})\exp(g_\theta(\mathbf{s}^{1:L},\mathbf{x}))\\
        &=\exp(g_\theta(\mathbf{s}^{1:L},\mathbf{x}))\prod_{l}p_{\text{LLM}}(\mathbf{s}^{l}|\mathbf{x},\mathbf{s}^{1:l-1}).
    \end{aligned}
\end{equation*}
To this end, various outputs generated by the black-box LLM are treated as distinct nodes.
The adapter then assigns scores to these nodes, thereby facilitating a heuristic selection of the most promising solution path that navigates through these sentence nodes.
For a beam size of $k$, at each step $l$, we generate $n$ samples of $\mathbf{s}^l$ based on $P_{\text{LLM}}(\mathbf{s}^l|\mathbf{x},\mathbf{s}^{1:l-1})$ for each beam. This results in $nk$ candidate chain hypotheses of $\mathbf{s}^{1:l}$, forming the candidate set $\mathcal{C}$. We then select the top-$k$ beams with the highest scores $g_\theta(\mathbf{s}^{1:l},\mathbf{x})$ given by the adapter, effectively pruning the beam options.
Once a pre-defined number of $L$ iterations is reached or all beams encounter a stop signal, we obtain $k$ reasoning steps. The adapted generation is then selected based on the highest-scoring option evaluated by the adapter.

\subsection{Online Adaptation}\label{subsec:method-oa}
According to the NCE loss function in Eq.(\ref{eq:final_loss}), it is essential to draw positive samples from the real distribution of the target domain, denoted as $\mathbf{y}_+\sim p_{\text{data}}(\mathbf{y}|\mathbf{x})$, and negative samples from its own generations, $\mathbf{y}_-\sim p_{\theta}(\mathbf{y}|\mathbf{x})$, to update the adapter parameters $\theta$.
However, an obvious disparity may arise between the real data distribution (\ie, the target domain) and its adapted generations (\ie, the source domain), resulting in overfitting to simplistic patterns and hindering the adapter from self-improvement.

We propose an online adaptation framework (Algorithm~\ref{alg:method}) with iterative sampling and training to address these challenges, drawing training samples from dynamic distributions. 
Initially, we establish and maintain separate sets for positive and negative samples. 
Then, for each iteration $t$, the online adaption framework involves three steps: 
(1) Sampling from the adapted inference $p_{\theta_t}(\mathbf{y}|\mathbf{x})$; (2) Updating the positive $\mathbf{y}^{(t)}_+$ and negative cases $\mathbf{y}^{(t)}_-$ based on feedback from human or AI; and (3) Updating the adapter parameters $\theta_{t+1}$ for the next iteration.

\begin{algorithm}[t]
    \caption{Overview of \method.}\label{alg:method}
    \label{algo:ada_cd}
    \begin{algorithmic}[1] % The number tells where the line numbering should start
    \STATE \textbf{Input:} $\Dcal=\{(\mathbf{x}_i,\mathbf{y}_i)\}_{i=1}^N$: Supervised fine-tuning dataset; $p_{\text{LLM}}$: Unadapted black-box LLM; $p_\theta$: Adapted LLM; $T$: Number of iterations; $\eta$: Learning rate; Beam size: $M$; \# Candidates generated per step: $K$.
        \STATE $p_\theta^{(0)}$ random initialization;
        \FOR{$t=0,\cdots, T-1$}
            \FOR{$i=1,\cdots,N$}
                \STATE Sample the candidates $\{\hat{\mathbf{y}}_{i,m}\}_{m=1}^M$ from the adapted inference via Eq.(\ref{eq:sample});
                \STATE Update the positive samples $\mathbf{y}_{i+}^{(t)}$ via Eq.(\ref{eq:pos});
                \STATE Update the negative samples $\mathbf{y}_{i-}^{(t)}$ via Eq.(\ref{eq:neg});
            \ENDFOR\\
            \STATE Compute $\nabla_\theta\ell(\theta_t)$ with $\mathbf{y}_{i+}^{(t)}$ and $\mathbf{y}_{i-}^{(t)}$ via Eq.\eqref{eq:final_loss};
            \STATE Update the adapter via Eq.(\ref{eq:para-up});
        \ENDFOR\\
    \textbf{Output:}  Fine-tuned $\theta_T$ after $T$-round iteration.
    \end{algorithmic}
\end{algorithm}

\textbf{Initialization.}
Prior to the iterative process, we establish two initial sets of positive and negative samples for adapter training.
Typically, positive samples are obtained from the ground-truth solutions, while negative samples are derived from the adapted inference $p_{\theta_0}$ by a randomly initialized adapter $\theta_0$.
In scenarios lacking ground-truth solutions, we alternatively employ human preferences for sourcing positive samples, or we utilize advanced LLMs (\eg, GPT-4) to generate AI feedback that closely aligns with human judgment~\cite{lee2023rlaif, bai2022constitutional, gilardi2023chatgpt}.
Mathematically, given each input query $\mathbf{x}_i$, we initially prompt a black-box LLM to generate $K$ responses $\{\mathbf{y}_{i,j}\}_{j=1}^{K}=\{\mathbf{y}_{i,1},\mathbf{y}_{i,2},\cdots,\mathbf{y}_{i,K}\}$. 
We then select the best response from the candidates as the positive sample, based on the ground-truth or human/AI feedback: $\mathbf{y}_{i+}^{(0)}=\mathbf{y}_{i,k}=\text{SEL}(\{\mathbf{y}_{i,j}\}_{j=1}^{K})$, where $k$ is the index of the best answer and $\text{SEL}(\cdot)$ indicates the selection according to feedback.
The rest candidates can then serve as negative cases: $\mathbf{y}_{i-}^{(0)}=\{\mathbf{y}_{i,j}|j\neq k\}_{j=1}^K$.

\noindent \textbf{Sampling from Adapted Inference.} 
To keep track of the dynamic distributions of $p_{\theta_t}$, at the beginning of each iteration $t$, we sample a set of $M$ candidates from the adapted inferences based on the current parameters $\theta_t$.
For each input sequence $\mathbf{x}_i$, we can sample the candidates:
\begin{equation}
    \begin{aligned}
         \{\hat{\mathbf{y}}_{i,m}\}_{m=1}^M\sim p_{\theta_{t}}(\mathbf{y}|\mathbf{x}_i).
    \end{aligned}\label{eq:sample}
\end{equation}

\noindent \textbf{Updating Training Data with Feedback.}
The initial positive set, comprising ground-truth solutions or preferred answers from advanced AI, may not be perfect and could contain some low-quality cases.
Moreover, the continuous learning of $\theta$ requires continual sampling from its own adapted inference as negative cases.
To accurately model the real data distribution $p_{\text{data}}$, 
we iteratively refine both the positive and negative training data by incorporating the previously sampled candidates from the adapted inference.
For each input sequence $\mathbf{x}_i$, we update the positive set by selecting a better answer from the previous positive samples $\mathbf{y}_-^{(t-1)}$ and the newly sampled candidates $\{\hat{\mathbf{y}}_{i,m}\}_{m=1}^M$ based on ground-truth or human/AI feedback:
\begin{equation}
    \begin{aligned}
        \mathbf{y}_{i+}^{(t)}&=\text{SEL}(\mathbf{y}_{i+}^{(t-1)},\{\hat{\mathbf{y}}_{i,m}\}_{m=1}^M).
    \end{aligned}\label{eq:pos}
\end{equation}
Subsequently, to ensure the selected positive answer is excluded from the candidate set, we update the negative samples with the remaining candidates:
\begin{equation}
    \begin{aligned}
        \mathbf{y}_{i-}^{(t)}&=\{\hat{\mathbf{y}}_{i,m}|\hat{\mathbf{y}}_{i,m}\neq\mathbf{y}_{i+}^{(t)}\}_{m=1}^M.
    \end{aligned}\label{eq:neg}
\end{equation}

\noindent \textbf{Update Adapter Parameters.}
With the updated positive samples $\mathbf{y}_+^{(t)}$ and negative samples $\mathbf{y}_-^{(t)}$, the last step of each iteration is to update the adapter parameters for the next iteration $\theta_{t+1}$.
By substituting the $\mathbf{y}_-$ and $\mathbf{y}_+$ in Eq.(\ref{eq:final_loss}), we can compute the gradient of loss function, $\nabla_\theta(\theta_t)$, and accordingly update the adapter parameters:
\begin{equation}
    \begin{aligned}
        \theta_{t+1}=\theta_{t}-\eta\nabla_\theta \ell(\theta_t),
    \end{aligned}\label{eq:para-up}
\end{equation}
where $\eta$ is the learning rate for the adapter update.

\section{Experiments}
In this section, we empirically examine the effectiveness of \method on black-box LLM adaptation to various tasks. 
We further analyze its flexibility (\ie, plug-and-play adaptation), cost-efficiency, ablations, scalability, and potential extensions for white-box LLM adaptation.

\subsection{Experiment Setup}

\textbf{Datasets.}
We evaluate \method on four distinct question-answering tasks, requiring model adaptation on mathematical (GSM8K~\cite{cobbe2021training}), implicit-reasoning (StrategyQA~\cite{geva-etal-2021-aristotle}), truthful (TruthfulQA~\cite{lin-etal-2022-truthfulqa}), and scientific (ScienceQA~\cite{lu2022scqa}) domains.
Dataset details are available in Appendix~\ref{app:dataset}.

\textbf{Baselines.}
We conduct our experiments using two base models for black-box adaptation: \texttt{gpt-3.5-turbo}~\cite{chatgpt} and \texttt{Mixtral-8$\times$7B}~\cite{jiang2024mixtral}.
We compare \method with the following baselines:
(1) \textbf{Chain-of-Thoughts (CoT)}~\cite{wei2022chain} represents the performance of the LLM without any adaptation. 
(2) \textbf{Supervised Fine-Tuning (SFT)} requires access to the base model's internal parameters and serves as the upper bound of the adaptation performance. 
For \texttt{gpt-3.5-turbo}, we use the OpenAI Fine-Tuning Service~\cite{SFTgpt} hosted on Azure~\cite{AzureSFT}.
For \texttt{Mixtral-8$\times$7B}, we contrast \method with the low-ranking adaptation (LoRA) under a SFT setting.
Additional baseline details can be found in Appendix~\ref{app:baseline}.

\begin{table*}[t]
\centering
\caption{Main results of adapting \texttt{gpt-3.5-turbo} on downstream tasks. For \method, we report the best performance of adapters with \# parameters of 0.1B and 0.3B. For all baselines and ours, we employ the CoT prompt as proposed in \cite{wei2022chain}.}\label{tab:main}
\fontsize{8.5}{10.5}\selectfont\setlength{\tabcolsep}{0.4em}
\begin{tabular}{@{}lc>{\columncolor{green!10}}cc>{\columncolor{green!10}}cc>{\columncolor{green!10}}cc>{\columncolor{green!10}}c@{}}
\toprule
\textbf{Dataset ($\rightarrow$)} & \multicolumn{2}{c}{\textbf{StrategyQA}} & \multicolumn{2}{c}{\textbf{GSM8K}} &  \multicolumn{2}{c}{\textbf{TruthfulQA}} & \multicolumn{2}{c}{\textbf{ScienceQA}}\\
\cmidrule(lr){2-3} \cmidrule(lr){4-5} \cmidrule(lr){6-7} \cmidrule{8-9}
\textbf{Adapter ($\downarrow$)} $\slash$ \textbf{Metrics ($\rightarrow$)} & Acc. (\%) & $\Delta$ (\%) & Acc. (\%) & $\Delta$ (\%) &  True + Info (\%) & $\Delta$ (\%) & Acc. (\%) & $\Delta$ (\%)  \\\midrule
\texttt{gpt-3.5-turbo}~\cite{chatgpt} & 66.59 & -       & 67.51         & -             & 77.00         & -         &  72.90       & -  \\
Azure-SFT~\cite{SFTgpt}               & 76.86 & +10.27  & 69.94         & +2.43         & 95.00         & +18.00    & 79.00    & +6.10  \\\midrule
\textbf{\method (Ground-Truth)}       & 71.62 & +5.03   & 73.86         & +6.35         & 79.70         & +2.70     &  78.53        & +5.63  \\
\textbf{\method  (AI Feedback)}       & 69.85 & +3.26   & 73.50         & +5.99         & 82.10         & +5.10     &  78.30        & +5.40 \\
\textbf{\method  (Combined)}          & \textbf{72.27}  & \textbf{+5.68} & \textbf{74.28} & \textbf{+6.77} & \textbf{83.60}         & \textbf{+6.60} & \textbf{79.40} & \textbf{+6.50} \\\bottomrule
\end{tabular}
\vspace{-2ex}
\end{table*}

\begin{table*}[t]
\centering
\caption{Results of plug-and-play adaptation on \texttt{davinci-002} and \texttt{Mixtral-8$\times$7B} across four datasets. For the plugger, we select \method tuned on \texttt{gpt-3.5-turbo} adaptation.}\label{tab:pnp}
\fontsize{8.5}{10.5}\selectfont\setlength{\tabcolsep}{0.4em}
\begin{tabular}{@{}lc>{\columncolor{green!10}}cc>{\columncolor{green!10}}cc>{\columncolor{green!10}}cc>{\columncolor{green!10}}c@{}}
\toprule
\textbf{Plugger ($\rightarrow$)} & \multicolumn{8}{c}{\textbf{\method (\texttt{gpt-3.5-turbo})}}\\\midrule
\textbf{Dataset ($\rightarrow$)} & \multicolumn{2}{c}{\textbf{StrategyQA}} & \multicolumn{2}{c}{\textbf{GSM8K}} &  \multicolumn{2}{c}{\textbf{TruthfulQA}} &\multicolumn{2}{c}{\textbf{Average}} \\
\cmidrule(lr){2-3} \cmidrule(lr){4-5} \cmidrule(lr){6-7} \cmidrule(lr){8-9}
\textbf{Black-Box LLMs ($\downarrow$)} $\slash$ \textbf{Metrics ($\rightarrow$)} & Acc. (\%) & $\Delta$ (\%) & Acc. (\%) & $\Delta$ (\%) &  True + Info (\%) & $\Delta$ (\%) & Acc. (\%) & $\Delta$ (\%)  \\\midrule
\texttt{davinci-002}                            & 44.19          & - & 23.73 & - & 31.50 & - & 33.14 & -   \\
\textbf{\texttt{davinci-002} (Plugged)}         & \textbf{59.61} & \textbf{+15.42} & \textbf{23.85} & \textbf{+0.12} & \textbf{36.50} & \textbf{+5.00} &  \textbf{39.99}  & \textbf{+6.85}  \\\midrule
\texttt{Mixtral-8$\times$7B}                    & 59.91          & - & 47.46 & - & 40.40 & - & 49.26 & -   \\
\textbf{\texttt{Mixtral-8$\times$7B} (Plugged)} & \textbf{63.97} & \textbf{+4.06}  & \textbf{47.61} & \textbf{+0.15} & \textbf{49.70} & \textbf{+9.30} & \textbf{53.76} &  \textbf{+4.50} \\\bottomrule
\end{tabular}
\vspace{-2ex}
\end{table*}

\textbf{Settings.}
To demonstrate the flexibility of our proposed method, we evaluate \method with three sources of labeled data: ground truth, AI feedback, and combined.
The settings are differentiated based on the source of positive sample selection:
(1) In the \textbf{Ground-Truth} setting, we utilize the ground-truth solutions originally provided by the dataset as positive samples, which remain constant throughout the entire online adaptation process.
(2) In the \textbf{AI Feedback} setting, we assume no access to any ground-truth information, neither step-wise solutions nor final answers. 
Following Section~\ref{subsec:method-oa}, we sample from the adapted inferences ($p_{\theta_t}$) to generate a set of candidates for each question.
An advanced LLM (\texttt{gpt-4}) is then used to simulate human preference, and the most preferred candidates are selected as positive samples. 
Detailed AI feedback selection criteria are available in Appendix \ref{app:aif}.
(3) In the \textbf{Combined} setting, the ground-truth set is augmented with preferred candidates obtained from the AI Feedback.
We also incorporate outcome supervision in all settings. We utilize the answers from the existing positive st to differentiate adapted inferences. Those inferences that align with the training set answers are treated as additional positive samples, while all others are considered negative.

\textbf{Implementations.}
For the \texttt{gpt-3.5-turbo}, we utilize the APIs provided by the Microsoft Azure OpenAI service. In the case of Mixtral-8$\times$7B, we employ the pre-trained checkpoint \texttt{mistralai/Mixtral-8x7B-v0.1} for model inference and parameter-efficient fine-tuning. 
Unless specified, \method employs \texttt{deberta-v3-base} (with 0.1B parameters) and \texttt{deberta-v3-large} (with 0.3B parameters) as backend models. The number of beams used for training and inference is set as 3 by default. Additional implementation details are available in Appendix~\ref{app:implementation1} and~\ref{app:implementation2}. The implementation of \method is available on GitHub\footnote{\href{https://github.com/haotiansun14/BBox-Adapter}{https://github.com/haotiansun14/BBox-Adapter}}.

\subsection{Main Results}\label{subsec:exp-main}
Table~\ref{tab:main} presents the main experimental results on three datasets under three distinct sources of positive samples.
\method consistently outperforms \texttt{gpt-3.5-turbo} by an average of $6.39\%$ across all datasets, highlighting its efficacy in adapting black-box LLMs to specific tasks.
Notably, \method (AI Feedback) demonstrates competitive performance compared to \method (Ground-Truth), which demonstrates its robust generalization capability across datasets, even in the absence of ground-truth answers.
Furthermore, \method (Combined) achieves the highest performance among the three variations.
This enhanced performance can be attributed to the combination of high-quality initial positive sets derived from ground-truth solutions and the dynamic updating of positive sets through AI feedback, leading to the continuous self-improvement of \method. 

\begin{table*}[t]
\centering
\fontsize{8.5}{10.5}\selectfont\setlength{\tabcolsep}{0.4em}
\caption{Comparison of performance and cost for the base model, SFT, and \method on the StrategyQA and GSM8K datasets. The performance is shown as accuracy (\%), while the costs (\$) are reported in training and inference expenses per thousand questions. Note that the inference cost was calculated by aggregating the total token consumption statistics provided by Azure API and subsequently applying the cost per token (\texttt{gpt-3.5-turbo-1106}) as specified in the OpenAI official documentation. The 'single step' refers to a simplified approach wherein the base model generates a set of complete answers in a single step and the adapter then selects the best answer from these candidates as the final response.}\label{tab:cost}
\begin{tabular}{@{}lc>{\columncolor{pink!10}}c>{\columncolor{blue!5}}cc>{\columncolor{pink!10}}c>{\columncolor{blue!5}}c@{}}
\toprule
\textbf{Dataset ($\rightarrow$)} & \multicolumn{3}{c}{\textbf{StrategyQA}} & \multicolumn{3}{c}{\textbf{GSM8K}} \\
\cmidrule(lr){2-4} \cmidrule(lr){5-7}
\textbf{Adapter ($\downarrow$) $\slash$ Metric ($\rightarrow$)}  & \textbf{Acc.(\%)} & \makecell{\textbf{Training}\\\textbf{Cost (\$)}} & \makecell{\textbf{Inference}\\ \textbf{Cost (\$)/1k Q}}& \textbf{Acc.(\%)} & \makecell{\textbf{Training}\\\textbf{Cost (\$)}} & \makecell{\textbf{Inference}\\ \textbf{Cost (\$)/1k Q}}\\\midrule
\texttt{gpt-3.5-turbo}       & 66.59 & -        & 0.41   &  67.51   & -         & 1.22  \\
Azure-SFT~\cite{SFTgpt}      & 76.86 & 153.00   & 7.50  & 69.94    & 216.50    & 28.30   \\\midrule
\method (Single-step)        & 69.87 & 2.77     & 2.20   & 71.13    & 7.54      & 3.10 \\
\method (Full-step)          & 71.62 & 3.48    &  5.37  & 74.28    & 11.58     & 12.46 \\\bottomrule
\end{tabular}
\vspace{-2.5ex}
\end{table*}

\subsection{Plug-and-Play Adaptation}\label{exp-pnp}
The tuned \method can be seamlessly applied to various black-box LLMs in a plug-and-play manner, eliminating the need for retraining or additional technical modifications. 
A well-trained version of \method adapting \texttt{gpt-3.5-turbo} can serve as a plugger to be integrated into the OpenAI base model \texttt{davinci-002} and \texttt{Mixtral-8$\times$7B}. Specifically, the adapter is employed to steer the generation processes of these models during the adapted inference of \method. 
Table~\ref{tab:pnp} presents the performance of \method on plug-and-play adaptation.
Compared to their unadapted black-box LLMs, \texttt{davinci-002} and \texttt{Mixtral-8$\times$7B}, our trained adapter demonstrates an average performance improvement of $6.85\%$ and $4.50\%$ across all three datasets, respectively. 
The effectiveness of \method in plug-and-play scenarios arises from its independence from the internal parameters of black-box LLMs. 
Unlike traditional SFT-related methods, which are generally inapplicable for plug-and-play adaptation due to their reliance on direct parameter manipulation, \method benefits from adapting text generation by analyzing data distributions.

\subsection{Cost Analysis}\label{subsec:exp-eff}
In Table~\ref{tab:cost}, we further compare the cost efficiency associated with different methods on the StrategyQA and GSM8K datasets.
Compared with the base model, Azure-SFT boosts accuracy by an average of $6.35\%$ at the expense of significantly higher costs.
\method, in single-step inference variant, brings $3.45\%$ performance gain compared with the base model, with $41.97$ times less training cost and $6.27$ times less inference cost than SFT.
Meanwhile, its full-step inference variant achieves $5.90\%$ improvement over the base model with $31.30$ times less training cost and $1.84$ times less inference cost.
This increased cost in its full-step variant is attributed to the integration of a beam search in the adapted inference, which requires the use of the black-box LLM APIs to generate multiple solution paths for selection.

\subsection{Ablation Study: Effect of Ranking-based NCE Loss}\label{subsec:exp-ablation}

\begin{table}[t]
\caption{Accuracy (\%) of \method fine-tuned with two types of loss: MLM loss and ranking-based NCE loss.}\label{tab:baseline_loss}
\centering
\fontsize{8}{10}\selectfont\setlength{\tabcolsep}{1.2em}
\begin{tabular}{@{}lcccc@{}}
\toprule
\textbf{Dataset ($\rightarrow$) }    & \multicolumn{2}{c}{\textbf{StrategyQA}} & \multicolumn{2}{c}{\textbf{GSM8K}}     \\ 
\textbf{Loss ($\downarrow$)}            & 0.1B          & 0.3B      & 0.1B        & 0.3B                            \\ \midrule
\textbf{MLM}                            & 61.52         &60.41      & 70.56       & 70.81 \\
\textbf{NCE} & \textbf{71.62}   & \textbf{71.18}    &\textbf{72.06} & \textbf{73.86} \\ \bottomrule
\end{tabular}
\vspace{-1ex}
\end{table}

We compare the efficacy of ranking-based NCE loss against the Masked Language Modeling (MLM) loss. For the MLM-based approach, we generate text chunks from the ground-truth data, randomly masking words, and then train the adapter using the masked word as supervision. During inference, we apply a similar process: masking a random word in each sequence generated by beam search and scoring the sequence based on the probability of the masked word.
The comparison results are detailed in Table~\ref{tab:baseline_loss}. 
\method with NCE loss consistently outperforms the baseline MLM loss approach, achieving improvements in task accuracy of up to 10\%. This demonstrates that the proposed loss effectively differentiates between the target and generated distributions and assigns scores accordingly.

\subsection{Scale Analysis}\label{subsec:exp-scale}
\vspace{-2mm}
We analyze the effect of scaling up \method by increasing the number of beams and iterations.

\begin{figure}[t]
	\centering
	\subfigure[Number of Beams]{
		\includegraphics[width=0.46\linewidth]{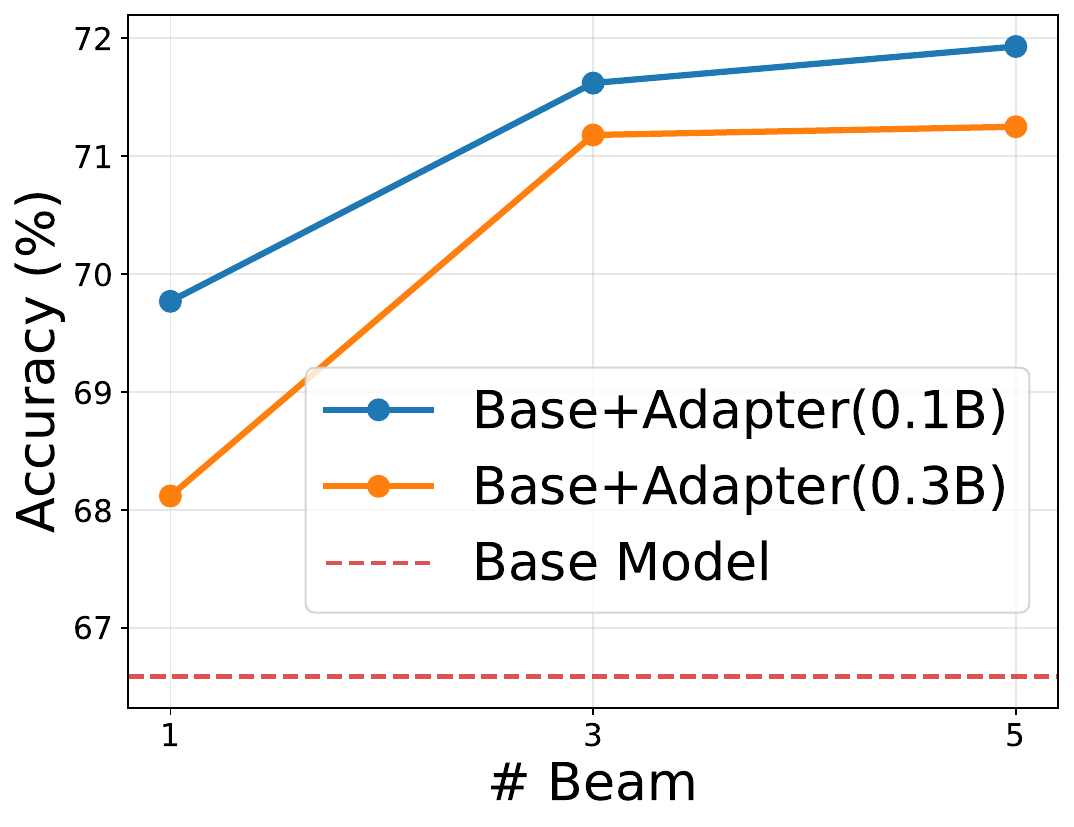}
		\label{fig:num-1}
	} 
     \subfigure[Number of Iterations]{
		\includegraphics[width=0.46\linewidth]{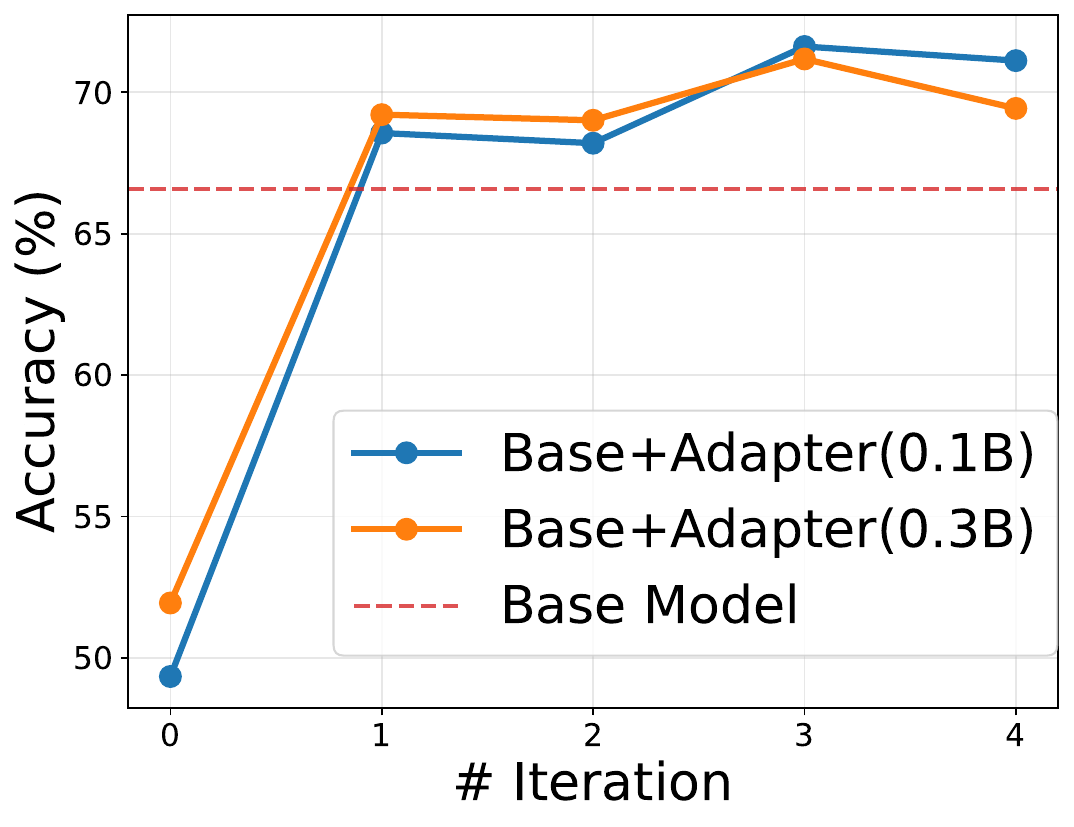}
		\label{fig:num-2}
	}
  \vspace{-2ex}
	\caption{Scale analysis on StrategyQA with (a) different beam sizes and (b) different iterations of online adaptation. Both experiments are conducted with two-shot prompting.}
 \vspace{-4ex}
\label{fig:num}
\end{figure}

\textbf{Number of Beams.}
We investigate three distinct beam sizes ($k=1,3,5$) within the context of \texttt{gpt-3.5-turbo} adaptation experiments on the StrategyQA dataset (Figure~\ref{fig:num-1}). 
Our results reveal that increasing the number of beams contributes to an average performance enhancement of $2.41\%$ across different adapter sizes (0.1B and 0.3B). 
The enhancement can likely be attributed to a larger beam retaining more candidate sequences at each decision step, thus expanding the search space. This broader search domain allows the black-box LLM to explore a wider variety of potential sequences, increasing the likelihood of identifying more optimal solutions for positive samples and improving the quantity and quality of negative cases.

\textbf{Number of Iterations.}
Figure~\ref{fig:num-2} presents the impact of different numbers of iterations ($T=0,1,2,3,4$) on model performance using the StrategyQA.
The un-finetuned adapter ($T=0$) performs even worse than the base model, which may assign inaccurate scores and misguide the beam search. 
The adapted LLM surpasses the performance of the base model after just one round of adaptation and shows consistent improvements with subsequent iterations, indicating the potential of \method for continuous self-improvement and task-specific refinement.
\vspace{-2mm}

\subsection{Extension on White-box Adaptation}
% \vspace{-2mm}
\begin{table}[t]
\centering
\caption{Accuracy (\%) and GPU memory usage on adapting \texttt{Mixtral-8$\times$7B} to the StrategyQA dataset.
VRAM refers to the maximum GPU memory required by each approach, where the base model (\texttt{Mixtral-8x7B}) is loaded in half-precision, and \method uses \texttt{BERT-0.1B} as the backend.}\label{tab:white}
\fontsize{8.5}{10.5}\selectfont\setlength{\tabcolsep}{0.4em}
\begin{tabular}{@{}lcccc@{}}
\toprule
\multirow{2}{*}{\textbf{Adapter ($\downarrow$) $\slash$ Metric ($\rightarrow$)} } & \multicolumn{2}{c}{\textbf{Acc.(\%)}} &\multicolumn{2}{c}{\textbf{VRAM (GiB)}}\\
                & 0.1B & 0.3B & Training & Inference\\\midrule
Base Model (\texttt{Mixtral-8x7B})     & \multicolumn{2}{c}{59.91} & - &     90    \\
Base + LoRA~\cite{hu2021lora}             & 73.80       & 75.98   &  208 & 92                       \\
Base + \method                & 66.08 & 65.26  & 105 & 92 \\\bottomrule
\end{tabular}
\vspace{-3ex}
\end{table}
\vspace{-2mm}
\begin{figure*}[!htb]
  \centering
  \includegraphics[width=0.9\linewidth]{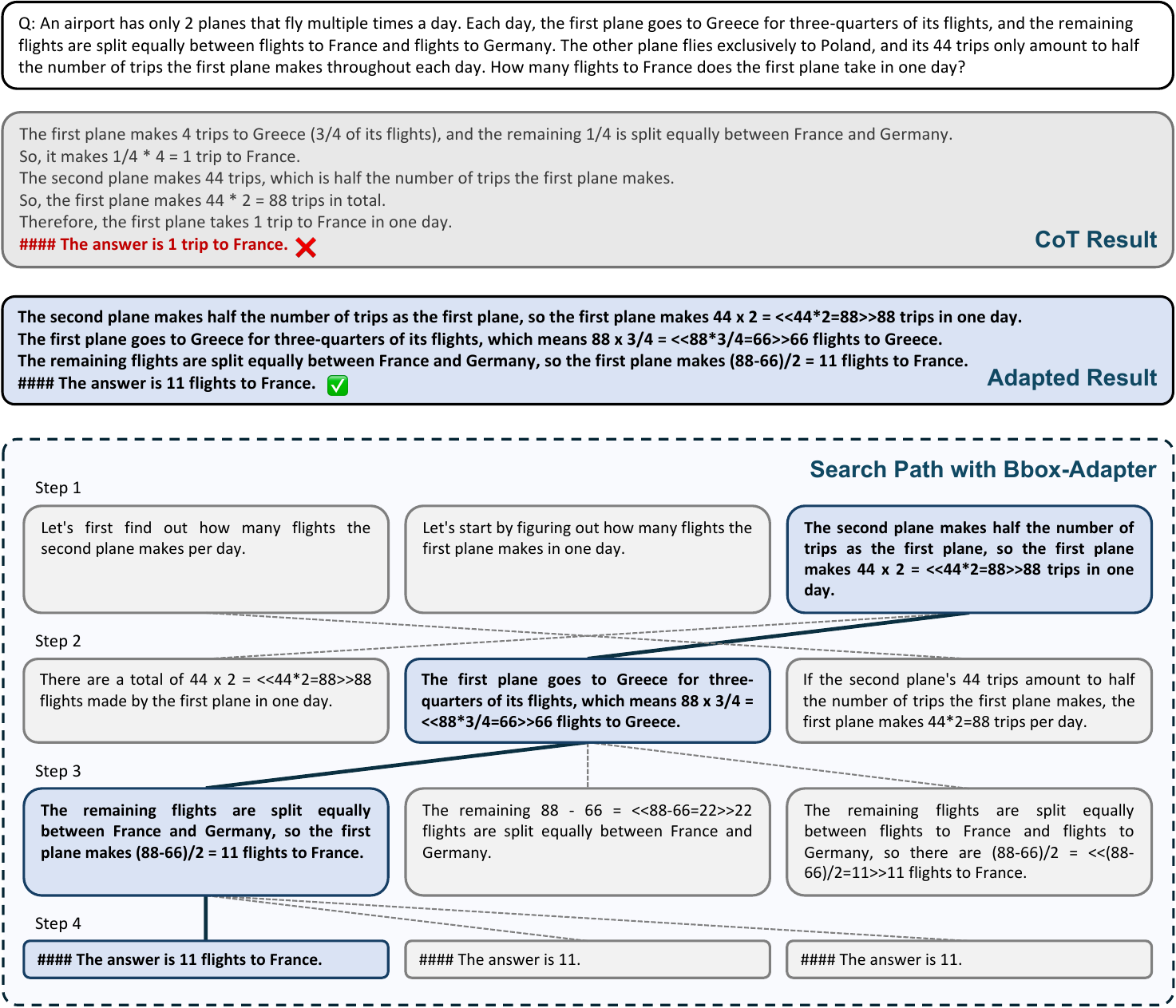}
  \caption{Case study of \method on GSM8K. For the given question, the CoT solution from original \texttt{gpt-3.5-turbo} is incorrect, while the model adapted using \method successfully executed a logical, step-by-step search, ultimately yielding the correct answer. For visualization, we display only top-3 candidate answers at each step.
  }
  \label{fig:case_studies}
  \vspace{-4mm}
\end{figure*}
We further extend the evaluation of \method to white-box LLMs, while treating them as black-box models (\ie, only using output generations without access to model parameters or output probabilities, therefore, preferable to the competitors). 
The results of adapting Mixtral-8$\times$7B in Table~\ref{tab:white} indicate that \method surpasses the base model (\texttt{Mixtral-8$\times$7B}) by 5.76\% on the StrategyQA dataset, demonstrating its strong reproducibility and generalization across different LMs. 
When comparing the adaptation of an equivalent number of parameters, SFT with the LoRA technique (SFT-LoRA) exhibits superior performance, due to its direct access to the model parameters.
In terms of resource utilization, \method requires less computational power and storage, making \method a more resource-efficient option for model adaptation.
\vspace{-1mm}
\subsection{Case Studies}
\vspace{-1mm}
Figure~\ref{fig:case_studies} presents a case study of \method applied to the GSM8K dataset. In this example, while the original \texttt{gpt-3.5-turbo} generates an incorrect answer to a given question, \method modified model successfully conducts a logical, step-by-step analysis, ultimately arriving at the correct solution.
\vspace{-2mm}
\subsection{Summary}
We summarize our main findings from empirical analysis as follows:
(1) \method significantly enhances the performance of base LLMs, demonstrating its effectiveness in adapting black-box LLMs without access to model parameters and output token probabilities.
(2) It exhibits flexibility irrespective of the availability of ground-truth solutions. Once fine-tuned by \method, the adapter seamlessly integrates with other black-box LLMs in a plug-and-play manner, eliminating the need for additional retraining.
(3) In comparison to SFT, \method achieves competitive performance at a significantly reduced cost.
\vspace{-2mm}

\vspace{-1mm}
\section{Conclusion}
\vspace{-1mm}
In this study, we presented \method, a novel and efficient approach for adapting black-box LLMs to specific tasks without requiring access to model parameters or output probabilities. 
By conceptualizing the adaptation process as a sampling problem within an EBM, \method effectively distinguishes between source and target domain data through a ranking-based NCE loss.
Extensive experiments demonstrate its effectiveness in adapting black-box LLMs to diverse tasks, enhancing model performance by up to $6.77\%$, and reducing training and inference costs by $31.30$x and $1.84$x, respectively.
\method addresses the challenges posed by the opaque nature of state-of-the-art LLMs, offering a transparent, privacy-conscious, and cost-effective solution for customizing black-box LLMs.

\section*{Acknowledgements}
This work was supported in part by NSF IIS-2008334, CAREER IIS-2144338, ONR MURI N00014-17-1-2656, and computing resources from Microsoft Azure.

\section*{Impact Statement}

\method addresses the challenges posed by the inherently opaque nature of state-of-the-art LLMs like GPT-4 and Bard, enabling the customization of black-box LLMs for personalized use cases. 
A key advantage of \method, compared to black-box LLM finetuning through API services, lies in its commitment to privacy through the fine-tuning of a smaller LM. It substantially reduces the privacy risks inherent in the transmission of confidential data to external APIs.
\method also stands out by eliminating the need for access to internal model weights or output probabilities, unlike existing white-box and grey-box adaptation methods. 
Fundamentally, \method can be interpreted as a natural way for adapting black-box LLMs to domain-specific tasks with transparency, privacy-consciousness, and cost-effectiveness.
\method holds considerable promise for positive social impact across diverse domains, including but not limited to customizing state-of-the-art black-box LLMs for enhancing personalized experience in privacy-sensitive applications. 

% A potential limitation of \method arises from its reliance on the quality of candidate answers generated by black-box LLMs. Although the proposed NCE loss aims to penalize negative samples and favor positive ones, its effectiveness may be compromised if these LLMs produce lower-quality outputs, potentially hindering \method's ability to select appropriate answers from these suboptimal candidates.
% For future work, we aim to incorporate knowledge retrieval framework into candidate generation to enhance output quality in domain-specific applications.

Given that \method is designed to reorient black-box Large Language Models (LLMs) from their initial source domain towards a designated target domain, there exists a non-negligible risk wherein individuals with malign intentions might engineer a detrimental target domain and accumulate injurious and toxic content for training purposes. While black-box LLMs inherently exhibit reluctance towards generating such content, our adapter could potentially be misappropriated to lure LLMs into producing such misguided outputs. Additionally, there is the conceivable risk that the gradient information from our proposed adapter, along with the logits bias inherent in black-box LLMs, could be exploited to orchestrate attacks or facilitate 'jailbreaking' in a manner akin to that described in prior works. We aim to mitigate these risks in our future studies.

% \section*{Acknowledgements}

% \textbf{Do not} include acknowledgements in the initial version of
% the paper submitted for blind review.

% If a paper is accepted, the final camera-ready version can (and
% probably should) include acknowledgements. In this case, please
% place such acknowledgements in an unnumbered section at the
% end of the paper. Typically, this will include thanks to reviewers
% who gave useful comments, to colleagues who contributed to the ideas,
% and to funding agencies and corporate sponsors that provided financial

\FloatBarrier

\bibliography{mybib}

\begin{thebibliography}{60}
\providecommand{\natexlab}[1]{#1}
\providecommand{\url}[1]{\texttt{#1}}
\expandafter\ifx\csname urlstyle\endcsname\relax
  \providecommand{\doi}[1]{doi: #1}\else
  \providecommand{\doi}{doi: \begingroup \urlstyle{rm}\Url}\fi

\bibitem[Anil et~al.(2023)Anil, Dai, Firat, Johnson, Lepikhin, Passos, Shakeri, Taropa, Bailey, Chen, et~al.]{anil2023palm2}
Anil, R., Dai, A.~M., Firat, O., Johnson, M., Lepikhin, D., Passos, A., Shakeri, S., Taropa, E., Bailey, P., Chen, Z., et~al.
\newblock Palm 2 technical report.
\newblock \emph{arXiv preprint arXiv:2305.10403}, 2023.

\bibitem[Bai et~al.(2022)Bai, Kadavath, Kundu, Askell, Kernion, Jones, Chen, Goldie, Mirhoseini, McKinnon, et~al.]{bai2022constitutional}
Bai, Y., Kadavath, S., Kundu, S., Askell, A., Kernion, J., Jones, A., Chen, A., Goldie, A., Mirhoseini, A., McKinnon, C., et~al.
\newblock Constitutional ai: Harmlessness from ai feedback.
\newblock \emph{arXiv preprint arXiv:2212.08073}, 2022.

\bibitem[Brown et~al.(2020)Brown, Mann, Ryder, Subbiah, Kaplan, Dhariwal, Neelakantan, Shyam, Sastry, Askell, et~al.]{brown2020language}
Brown, T., Mann, B., Ryder, N., Subbiah, M., Kaplan, J.~D., Dhariwal, P., Neelakantan, A., Shyam, P., Sastry, G., Askell, A., et~al.
\newblock Language models are few-shot learners.
\newblock \emph{Advances in neural information processing systems}, 33:\penalty0 1877--1901, 2020.

\bibitem[Chowdhery et~al.(2022)Chowdhery, Narang, Devlin, Bosma, Mishra, Roberts, Barham, Chung, Sutton, Gehrmann, et~al.]{chowdhery2022palm}
Chowdhery, A., Narang, S., Devlin, J., Bosma, M., Mishra, G., Roberts, A., Barham, P., Chung, H.~W., Sutton, C., Gehrmann, S., et~al.
\newblock Palm: Scaling language modeling with pathways.
\newblock \emph{arXiv preprint arXiv:2204.02311}, 2022.

\bibitem[Chung et~al.(2022)Chung, Hou, Longpre, Zoph, Tay, Fedus, Li, Wang, Dehghani, Brahma, et~al.]{chung2022scaling}
Chung, H.~W., Hou, L., Longpre, S., Zoph, B., Tay, Y., Fedus, W., Li, Y., Wang, X., Dehghani, M., Brahma, S., et~al.
\newblock Scaling instruction-finetuned language models.
\newblock \emph{arXiv preprint arXiv:2210.11416}, 2022.

\bibitem[Cobbe et~al.(2021)Cobbe, Kosaraju, Bavarian, Chen, Jun, Kaiser, Plappert, Tworek, Hilton, Nakano, et~al.]{cobbe2021training}
Cobbe, K., Kosaraju, V., Bavarian, M., Chen, M., Jun, H., Kaiser, L., Plappert, M., Tworek, J., Hilton, J., Nakano, R., et~al.
\newblock Training verifiers to solve math word problems.
\newblock \emph{arXiv preprint arXiv:2110.14168}, 2021.

\bibitem[Deng et~al.(2020)Deng, Bakhtin, Ott, Szlam, and Ranzato]{deng2020residual}
Deng, Y., Bakhtin, A., Ott, M., Szlam, A., and Ranzato, M.
\newblock Residual energy-based models for text generation.
\newblock \emph{arXiv preprint arXiv:2004.11714}, 2020.

\bibitem[Devlin et~al.(2019)Devlin, Chang, Lee, and Toutanova]{devlin-etal-2019-bert}
Devlin, J., Chang, M.-W., Lee, K., and Toutanova, K.
\newblock {BERT}: Pre-training of deep bidirectional transformers for language understanding.
\newblock In Burstein, J., Doran, C., and Solorio, T. (eds.), \emph{Proceedings of the 2019 Conference of the North {A}merican Chapter of the Association for Computational Linguistics: Human Language Technologies, Volume 1 (Long and Short Papers)}, pp.\  4171--4186, Minneapolis, Minnesota, June 2019. Association for Computational Linguistics.
\newblock \doi{10.18653/v1/N19-1423}.

\bibitem[Du \& Mordatch(2019)Du and Mordatch]{du2019implicit_ebm}
Du, Y. and Mordatch, I.
\newblock Implicit generation and generalization in energy-based models.
\newblock \emph{arXiv preprint arXiv:1903.08689}, 2019.

\bibitem[Geva et~al.(2021)Geva, Khashabi, Segal, Khot, Roth, and Berant]{geva-etal-2021-aristotle}
Geva, M., Khashabi, D., Segal, E., Khot, T., Roth, D., and Berant, J.
\newblock Did aristotle use a laptop? a question answering benchmark with implicit reasoning strategies.
\newblock \emph{Transactions of the Association for Computational Linguistics}, 9:\penalty0 346--361, 2021.
\newblock \doi{10.1162/tacl_a_00370}.

\bibitem[Gilardi et~al.(2023)Gilardi, Alizadeh, and Kubli]{gilardi2023chatgpt}
Gilardi, F., Alizadeh, M., and Kubli, M.
\newblock Chatgpt outperforms crowd workers for text-annotation tasks.
\newblock \emph{Proceedings of the National Academy of Sciences}, 120\penalty0 (30):\penalty0 e2305016120, 2023.
\newblock \doi{10.1073/pnas.2305016120}.

\bibitem[Golovneva et~al.(2023)Golovneva, O'Brien, Pasunuru, Wang, Zettlemoyer, Fazel-Zarandi, and Celikyilmaz]{golovneva2023pathfinder}
Golovneva, O., O'Brien, S., Pasunuru, R., Wang, T., Zettlemoyer, L., Fazel-Zarandi, M., and Celikyilmaz, A.
\newblock Pathfinder: Guided search over multi-step reasoning paths.
\newblock \emph{arXiv preprint arXiv:2312.05180}, 2023.

\bibitem[Gupta et~al.(2023)Gupta, Th{\'e}rien, Ibrahim, Richter, Anthony, Belilovsky, Rish, and Lesort]{gupta2023continual}
Gupta, K., Th{\'e}rien, B., Ibrahim, A., Richter, M.~L., Anthony, Q.~G., Belilovsky, E., Rish, I., and Lesort, T.
\newblock Continual pre-training of large language models: How to re-warm your model?
\newblock In \emph{Workshop on Efficient Systems for Foundation Models@ ICML2023}, 2023.

\bibitem[Gururangan et~al.(2020)Gururangan, Marasovi{\'c}, Swayamdipta, Lo, Beltagy, Downey, and Smith]{gururangan-etal-2020-dont}
Gururangan, S., Marasovi{\'c}, A., Swayamdipta, S., Lo, K., Beltagy, I., Downey, D., and Smith, N.~A.
\newblock Don{'}t stop pretraining: Adapt language models to domains and tasks.
\newblock In Jurafsky, D., Chai, J., Schluter, N., and Tetreault, J. (eds.), \emph{Proceedings of the 58th Annual Meeting of the Association for Computational Linguistics}, pp.\  8342--8360, Online, July 2020. Association for Computational Linguistics.
\newblock \doi{10.18653/v1/2020.acl-main.740}.

\bibitem[Gutmann \& Hyv{\"a}rinen(2010)Gutmann and Hyv{\"a}rinen]{gutmann2010noise}
Gutmann, M. and Hyv{\"a}rinen, A.
\newblock Noise-contrastive estimation: A new estimation principle for unnormalized statistical models.
\newblock In \emph{Proceedings of the thirteenth international conference on artificial intelligence and statistics}, pp.\  297--304. JMLR Workshop and Conference Proceedings, 2010.

\bibitem[Hao et~al.(2023)Hao, Gu, Ma, Hong, Wang, Wang, and Hu]{hao-etal-2023-reasoning}
Hao, S., Gu, Y., Ma, H., Hong, J., Wang, Z., Wang, D., and Hu, Z.
\newblock Reasoning with language model is planning with world model.
\newblock In Bouamor, H., Pino, J., and Bali, K. (eds.), \emph{Proceedings of the 2023 Conference on Empirical Methods in Natural Language Processing}, pp.\  8154--8173, Singapore, December 2023. Association for Computational Linguistics.
\newblock \doi{10.18653/v1/2023.emnlp-main.507}.

\bibitem[Hartvigsen et~al.(2022)Hartvigsen, Gabriel, Palangi, Sap, Ray, and Kamar]{hartvigsen2022toxigen}
Hartvigsen, T., Gabriel, S., Palangi, H., Sap, M., Ray, D., and Kamar, E.
\newblock Toxigen: A large-scale machine-generated dataset for adversarial and implicit hate speech detection.
\newblock \emph{arXiv preprint arXiv:2203.09509}, 2022.

\bibitem[He et~al.(2021)He, Zhou, Ma, Berg-Kirkpatrick, and Neubig]{he2021towards}
He, J., Zhou, C., Ma, X., Berg-Kirkpatrick, T., and Neubig, G.
\newblock Towards a unified view of parameter-efficient transfer learning.
\newblock In \emph{International Conference on Learning Representations}, 2021.

\bibitem[Houlsby et~al.(2019)Houlsby, Giurgiu, Jastrzebski, Morrone, De~Laroussilhe, Gesmundo, Attariyan, and Gelly]{houlsby2019parameter}
Houlsby, N., Giurgiu, A., Jastrzebski, S., Morrone, B., De~Laroussilhe, Q., Gesmundo, A., Attariyan, M., and Gelly, S.
\newblock Parameter-efficient transfer learning for nlp.
\newblock In \emph{International Conference on Machine Learning}, pp.\  2790--2799. PMLR, 2019.

\bibitem[Hu et~al.(2021)Hu, Wallis, Allen-Zhu, Li, Wang, Wang, Chen, et~al.]{hu2021lora}
Hu, E.~J., Wallis, P., Allen-Zhu, Z., Li, Y., Wang, S., Wang, L., Chen, W., et~al.
\newblock Lora: Low-rank adaptation of large language models.
\newblock In \emph{International Conference on Learning Representations}, 2021.

\bibitem[Hu et~al.(2023)Hu, Mitchell, Manning, and Finn]{hu2023meta}
Hu, N., Mitchell, E., Manning, C., and Finn, C.
\newblock Meta-learning online adaptation of language models.
\newblock In \emph{Proceedings of the 2023 Conference on Empirical Methods in Natural Language Processing}, pp.\  4418--4432, Singapore, December 2023. Association for Computational Linguistics.

\bibitem[Huang et~al.(2023)Huang, Liu, Zhong, Shi, and Lee]{huang2023k}
Huang, Y., Liu, D., Zhong, Z., Shi, W., and Lee, Y.~T.
\newblock $k$ nn-adapter: Efficient domain adaptation for black-box language models.
\newblock \emph{arXiv preprint arXiv:2302.10879}, 2023.

\bibitem[Jiang et~al.(2024)Jiang, Sablayrolles, Roux, Mensch, Savary, Bamford, Chaplot, Casas, Hanna, Bressand, et~al.]{jiang2024mixtral}
Jiang, A.~Q., Sablayrolles, A., Roux, A., Mensch, A., Savary, B., Bamford, C., Chaplot, D.~S., Casas, D. d.~l., Hanna, E.~B., Bressand, F., et~al.
\newblock Mixtral of experts.
\newblock \emph{arXiv preprint arXiv:2401.04088}, 2024.

\bibitem[Kadavath et~al.(2022)Kadavath, Conerly, Askell, Henighan, Drain, Perez, Schiefer, Hatfield-Dodds, DasSarma, Tran-Johnson, et~al.]{kadavath2022language}
Kadavath, S., Conerly, T., Askell, A., Henighan, T., Drain, D., Perez, E., Schiefer, N., Hatfield-Dodds, Z., DasSarma, N., Tran-Johnson, E., et~al.
\newblock Language models (mostly) know what they know.
\newblock \emph{arXiv preprint arXiv:2207.05221}, 2022.

\bibitem[Ke et~al.(2022)Ke, Shao, Lin, Konishi, Kim, and Liu]{ke2022continual}
Ke, Z., Shao, Y., Lin, H., Konishi, T., Kim, G., and Liu, B.
\newblock Continual pre-training of language models.
\newblock In \emph{The Eleventh International Conference on Learning Representations}, 2022.

\bibitem[Khalifa et~al.(2023)Khalifa, Logeswaran, Lee, Lee, and Wang]{khalifa2023grace}
Khalifa, M., Logeswaran, L., Lee, M., Lee, H., and Wang, L.
\newblock Grace: Discriminator-guided chain-of-thought reasoning, 2023.

\bibitem[Lee et~al.(2023)Lee, Phatale, Mansoor, Lu, Mesnard, Bishop, Carbune, and Rastogi]{lee2023rlaif}
Lee, H., Phatale, S., Mansoor, H., Lu, K., Mesnard, T., Bishop, C., Carbune, V., and Rastogi, A.
\newblock Rlaif: Scaling reinforcement learning from human feedback with ai feedback.
\newblock \emph{arXiv preprint arXiv:2309.00267}, 2023.

\bibitem[Li \& Liang(2021)Li and Liang]{li2021prefix}
Li, X.~L. and Liang, P.
\newblock Prefix-tuning: Optimizing continuous prompts for generation.
\newblock In \emph{Proceedings of the 59th Annual Meeting of the Association for Computational Linguistics and the 11th International Joint Conference on Natural Language Processing (Volume 1: Long Papers)}, pp.\  4582--4597, 2021.

\bibitem[Li et~al.(2023)Li, Lin, Zhang, Fu, Chen, Lou, and Chen]{li-etal-2023-making}
Li, Y., Lin, Z., Zhang, S., Fu, Q., Chen, B., Lou, J.-G., and Chen, W.
\newblock Making language models better reasoners with step-aware verifier.
\newblock In Rogers, A., Boyd-Graber, J., and Okazaki, N. (eds.), \emph{Proceedings of the 61st Annual Meeting of the Association for Computational Linguistics (Volume 1: Long Papers)}, pp.\  5315--5333, Toronto, Canada, July 2023. Association for Computational Linguistics.
\newblock \doi{10.18653/v1/2023.acl-long.291}.

\bibitem[Lin et~al.(2022)Lin, Hilton, and Evans]{lin-etal-2022-truthfulqa}
Lin, S., Hilton, J., and Evans, O.
\newblock {T}ruthful{QA}: Measuring how models mimic human falsehoods.
\newblock In Muresan, S., Nakov, P., and Villavicencio, A. (eds.), \emph{Proceedings of the 60th Annual Meeting of the Association for Computational Linguistics (Volume 1: Long Papers)}, pp.\  3214--3252, Dublin, Ireland, May 2022. Association for Computational Linguistics.
\newblock \doi{10.18653/v1/2022.acl-long.229}.

\bibitem[Liu et~al.(2024)Liu, Han, Wang, Tsvetkov, Choi, and Smith]{liu2024tuning}
Liu, A., Han, X., Wang, Y., Tsvetkov, Y., Choi, Y., and Smith, N.~A.
\newblock Tuning language models by proxy, 2024.

\bibitem[Liu et~al.(2022)Liu, Ji, Fu, Tam, Du, Yang, and Tang]{liu-etal-2022-p}
Liu, X., Ji, K., Fu, Y., Tam, W., Du, Z., Yang, Z., and Tang, J.
\newblock {P}-tuning: Prompt tuning can be comparable to fine-tuning across scales and tasks.
\newblock In Muresan, S., Nakov, P., and Villavicencio, A. (eds.), \emph{Proceedings of the 60th Annual Meeting of the Association for Computational Linguistics (Volume 2: Short Papers)}, pp.\  61--68, Dublin, Ireland, May 2022. Association for Computational Linguistics.
\newblock \doi{10.18653/v1/2022.acl-short.8}.

\bibitem[Lu et~al.(2022)Lu, Mishra, Xia, Qiu, Chang, Zhu, Tafjord, Clark, and Kalyan]{lu2022scqa}
Lu, P., Mishra, S., Xia, T., Qiu, L., Chang, K.-W., Zhu, S.-C., Tafjord, O., Clark, P., and Kalyan, A.
\newblock Learn to explain: Multimodal reasoning via thought chains for science question answering, 2022.

\bibitem[Lu et~al.(2023)Lu, Brahman, West, Jung, Chandu, Ravichander, Ammanabrolu, Jiang, Ramnath, Dziri, et~al.]{lu2023ipa}
Lu, X., Brahman, F., West, P., Jung, J., Chandu, K., Ravichander, A., Ammanabrolu, P., Jiang, L., Ramnath, S., Dziri, N., et~al.
\newblock Inference-time policy adapters (ipa): Tailoring extreme-scale lms without fine-tuning.
\newblock In \emph{Proceedings of the 2023 Conference on Empirical Methods in Natural Language Processing}, pp.\  6863--6883, 2023.

\bibitem[Ma \& Collins(2018)Ma and Collins]{ma2018noise}
Ma, Z. and Collins, M.
\newblock Noise contrastive estimation and negative sampling for conditional models: Consistency and statistical efficiency.
\newblock In Riloff, E., Chiang, D., Hockenmaier, J., and Tsujii, J. (eds.), \emph{Proceedings of the 2018 Conference on Empirical Methods in Natural Language Processing}, pp.\  3698--3707, Brussels, Belgium, October-November 2018. Association for Computational Linguistics.
\newblock \doi{10.18653/v1/D18-1405}.

\bibitem[Madaan et~al.(2023)Madaan, Tandon, Gupta, Hallinan, Gao, Wiegreffe, Alon, Dziri, Prabhumoye, Yang, et~al.]{madaan2023self}
Madaan, A., Tandon, N., Gupta, P., Hallinan, S., Gao, L., Wiegreffe, S., Alon, U., Dziri, N., Prabhumoye, S., Yang, Y., et~al.
\newblock Self-refine: Iterative refinement with self-feedback.
\newblock \emph{arXiv preprint arXiv:2303.17651}, 2023.

\bibitem[Microsoft(2023)]{AzureSFT}
Microsoft.
\newblock Azure openai gpt 3.5 turbo fine-tuning tutorial.
\newblock \emph{Microsoft Learn Tutorial}, 2023.

\bibitem[Oord et~al.(2018)Oord, Li, and Vinyals]{oord2018representation}
Oord, A. v.~d., Li, Y., and Vinyals, O.
\newblock Representation learning with contrastive predictive coding.
\newblock \emph{arXiv preprint arXiv:1807.03748}, 2018.

\bibitem[OpenAI(2022)]{chatgpt}
OpenAI.
\newblock Introducing chatgpt.
\newblock \emph{OpenAI Blog}, 2022.
\newblock URL \url{https://openai.com/blog/chatgpt}.

\bibitem[OpenAI(2023)]{openai2023}
OpenAI.
\newblock Gpt-4 technical report.
\newblock \emph{arXiv}, pp.\  2303.08774v3, 2023.

\bibitem[Ormazabal et~al.(2023)Ormazabal, Artetxe, and Agirre]{ormazabal-etal-2023-comblm}
Ormazabal, A., Artetxe, M., and Agirre, E.
\newblock {C}omb{LM}: Adapting black-box language models through small fine-tuned models.
\newblock In Bouamor, H., Pino, J., and Bali, K. (eds.), \emph{Proceedings of the 2023 Conference on Empirical Methods in Natural Language Processing}, pp.\  2961--2974, Singapore, December 2023. Association for Computational Linguistics.
\newblock \doi{10.18653/v1/2023.emnlp-main.180}.

\bibitem[Paul et~al.(2023)Paul, Ismayilzada, Peyrard, Borges, Bosselut, West, and Faltings]{paul2023refiner}
Paul, D., Ismayilzada, M., Peyrard, M., Borges, B., Bosselut, A., West, R., and Faltings, B.
\newblock Refiner: Reasoning feedback on intermediate representations.
\newblock \emph{arXiv preprint arXiv:2304.01904}, 2023.

\bibitem[Peng et~al.(2023)Peng, Wu, Allard, Kilpatrick, and Heidel]{SFTgpt}
Peng, A., Wu, M., Allard, J., Kilpatrick, L., and Heidel, S.
\newblock Gpt-3.5 turbo fine-tuning and api updates.
\newblock \emph{OpenAI Blog}, 2023.
\newblock URL \url{https://openai.com/blog/gpt-3-5-turbo-fine-tuning-and-api-updates}.

\bibitem[Radford et~al.(2018)Radford, Narasimhan, Salimans, and Sutskever]{radfordimproving}
Radford, A., Narasimhan, K., Salimans, T., and Sutskever, I.
\newblock Improving language understanding by generative pre-training.
\newblock \emph{OpenAI Blog}, 2018.

\bibitem[Radford et~al.(2019)Radford, Wu, Child, Luan, Amodei, and Sutskever]{radfordlanguage}
Radford, A., Wu, J., Child, R., Luan, D., Amodei, D., and Sutskever, I.
\newblock Language models are unsupervised multitask learners.
\newblock \emph{OpenAI Blog}, 2019.

\bibitem[Shinn et~al.(2023)Shinn, Cassano, Gopinath, Narasimhan, and Yao]{shinn2023reflexion}
Shinn, N., Cassano, F., Gopinath, A., Narasimhan, K.~R., and Yao, S.
\newblock Reflexion: Language agents with verbal reinforcement learning.
\newblock In \emph{Thirty-seventh Conference on Neural Information Processing Systems}, 2023.

\bibitem[Sun et~al.(2022)Sun, Shao, Qian, Huang, and Qiu]{sun2022black}
Sun, T., Shao, Y., Qian, H., Huang, X., and Qiu, X.
\newblock Black-box tuning for language-model-as-a-service.
\newblock In \emph{International Conference on Machine Learning}, pp.\  20841--20855. PMLR, 2022.

\bibitem[Team et~al.(2023)Team, Anil, Borgeaud, Wu, Alayrac, Yu, Soricut, Schalkwyk, Dai, Hauth, et~al.]{team2023gemini}
Team, G., Anil, R., Borgeaud, S., Wu, Y., Alayrac, J.-B., Yu, J., Soricut, R., Schalkwyk, J., Dai, A.~M., Hauth, A., et~al.
\newblock Gemini: a family of highly capable multimodal models.
\newblock \emph{arXiv preprint arXiv:2312.11805}, 2023.

\bibitem[Touvron et~al.(2023)Touvron, Martin, Stone, Albert, Almahairi, Babaei, Bashlykov, Batra, Bhargava, Bhosale, et~al.]{touvron2023llama}
Touvron, H., Martin, L., Stone, K., Albert, P., Almahairi, A., Babaei, Y., Bashlykov, N., Batra, S., Bhargava, P., Bhosale, S., et~al.
\newblock Llama 2: Open foundation and fine-tuned chat models.
\newblock \emph{arXiv preprint arXiv:2307.09288}, 2023.

\bibitem[Wang et~al.(2023{\natexlab{a}})Wang, Li, Chen, Song, Lin, Cao, Liu, and Sui]{wang2023making}
Wang, P., Li, L., Chen, L., Song, F., Lin, B., Cao, Y., Liu, T., and Sui, Z.
\newblock Making large language models better reasoners with alignment.
\newblock \emph{arXiv preprint arXiv:2309.02144}, 2023{\natexlab{a}}.

\bibitem[Wang et~al.(2023{\natexlab{b}})Wang, Li, Shao, Xu, Dai, Li, Chen, Wu, and Sui]{wang2023math}
Wang, P., Li, L., Shao, Z., Xu, R., Dai, D., Li, Y., Chen, D., Wu, Y., and Sui, Z.
\newblock Math-shepherd: A label-free step-by-step verifier for llms in mathematical reasoning.
\newblock \emph{arXiv preprint arXiv:2312.08935}, 2023{\natexlab{b}}.

\bibitem[Wang et~al.(2022{\natexlab{a}})Wang, Wei, Schuurmans, Le, Chi, Narang, Chowdhery, and Zhou]{wang2022self}
Wang, X., Wei, J., Schuurmans, D., Le, Q.~V., Chi, E.~H., Narang, S., Chowdhery, A., and Zhou, D.
\newblock Self-consistency improves chain of thought reasoning in language models.
\newblock In \emph{The Eleventh International Conference on Learning Representations}, 2022{\natexlab{a}}.

\bibitem[Wang et~al.(2022{\natexlab{b}})Wang, Mishra, Alipoormolabashi, Kordi, Mirzaei, Naik, Ashok, Dhanasekaran, Arunkumar, Stap, Pathak, Karamanolakis, Lai, Purohit, Mondal, Anderson, Kuznia, Doshi, Pal, Patel, Moradshahi, Parmar, Purohit, Varshney, Kaza, Verma, Puri, Karia, Doshi, Sampat, Mishra, Reddy~A, Patro, Dixit, and Shen]{wang-etal-2022-super}
Wang, Y., Mishra, S., Alipoormolabashi, P., Kordi, Y., Mirzaei, A., Naik, A., Ashok, A., Dhanasekaran, A.~S., Arunkumar, A., Stap, D., Pathak, E., Karamanolakis, G., Lai, H., Purohit, I., Mondal, I., Anderson, J., Kuznia, K., Doshi, K., Pal, K.~K., Patel, M., Moradshahi, M., Parmar, M., Purohit, M., Varshney, N., Kaza, P.~R., Verma, P., Puri, R.~S., Karia, R., Doshi, S., Sampat, S.~K., Mishra, S., Reddy~A, S., Patro, S., Dixit, T., and Shen, X.
\newblock Super-{N}atural{I}nstructions: Generalization via declarative instructions on 1600+ {NLP} tasks.
\newblock In Goldberg, Y., Kozareva, Z., and Zhang, Y. (eds.), \emph{Proceedings of the 2022 Conference on Empirical Methods in Natural Language Processing}, pp.\  5085--5109, Abu Dhabi, United Arab Emirates, December 2022{\natexlab{b}}. Association for Computational Linguistics.
\newblock \doi{10.18653/v1/2022.emnlp-main.340}.

\bibitem[Wei et~al.(2021)Wei, Bosma, Zhao, Guu, Yu, Lester, Du, Dai, and Le]{wei2021finetuned}
Wei, J., Bosma, M., Zhao, V., Guu, K., Yu, A.~W., Lester, B., Du, N., Dai, A.~M., and Le, Q.~V.
\newblock Finetuned language models are zero-shot learners.
\newblock In \emph{International Conference on Learning Representations}, 2021.

\bibitem[Wei et~al.(2022)Wei, Wang, Schuurmans, Bosma, Xia, Chi, Le, Zhou, et~al.]{wei2022chain}
Wei, J., Wang, X., Schuurmans, D., Bosma, M., Xia, F., Chi, E., Le, Q.~V., Zhou, D., et~al.
\newblock Chain-of-thought prompting elicits reasoning in large language models.
\newblock \emph{Advances in Neural Information Processing Systems}, 35:\penalty0 24824--24837, 2022.

\bibitem[Xie et~al.(2023)Xie, Kawaguchi, Zhao, Zhao, Kan, He, and Xie]{xie2023self}
Xie, Y., Kawaguchi, K., Zhao, Y., Zhao, X., Kan, M.-Y., He, J., and Xie, Q.
\newblock Self-evaluation guided beam search for reasoning.
\newblock In \emph{Thirty-seventh Conference on Neural Information Processing Systems}, 2023.

\bibitem[Yao et~al.(2023)Yao, Yu, Zhao, Shafran, Griffiths, Cao, and Narasimhan]{yao2023tree}
Yao, S., Yu, D., Zhao, J., Shafran, I., Griffiths, T.~L., Cao, Y., and Narasimhan, K.~R.
\newblock Tree of thoughts: Deliberate problem solving with large language models.
\newblock In \emph{Thirty-seventh Conference on Neural Information Processing Systems}, 2023.

\bibitem[Zhou et~al.(2022)Zhou, Sch{\"a}rli, Hou, Wei, Scales, Wang, Schuurmans, Cui, Bousquet, Le, et~al.]{zhou2022least}
Zhou, D., Sch{\"a}rli, N., Hou, L., Wei, J., Scales, N., Wang, X., Schuurmans, D., Cui, C., Bousquet, O., Le, Q.~V., et~al.
\newblock Least-to-most prompting enables complex reasoning in large language models.
\newblock In \emph{The Eleventh International Conference on Learning Representations}, 2022.

\bibitem[Zhu et~al.(2023)Zhu, Wang, Zhang, Zhang, Huang, Gan, Zhang, and Yang]{zhu-etal-2023-solving}
Zhu, X., Wang, J., Zhang, L., Zhang, Y., Huang, Y., Gan, R., Zhang, J., and Yang, Y.
\newblock Solving math word problems via cooperative reasoning induced language models.
\newblock In Rogers, A., Boyd-Graber, J., and Okazaki, N. (eds.), \emph{Proceedings of the 61st Annual Meeting of the Association for Computational Linguistics (Volume 1: Long Papers)}, pp.\  4471--4485, Toronto, Canada, July 2023. Association for Computational Linguistics.
\newblock \doi{10.18653/v1/2023.acl-long.245}.

\bibitem[Zhuang et~al.(2023)Zhuang, Chen, Yu, Mitra, Bursztyn, Rossi, Sarkhel, and Zhang]{zhuang2023toolchain}
Zhuang, Y., Chen, X., Yu, T., Mitra, S., Bursztyn, V., Rossi, R.~A., Sarkhel, S., and Zhang, C.
\newblock Toolchain*: Efficient action space navigation in large language models with a* search.
\newblock \emph{arXiv preprint arXiv:2310.13227}, 2023.

\end{thebibliography}
\bibliographystyle{icml2024}

%%%%%%%%%%%%%%%%%%%%%%%%%%%%%%%%%%%%%%%%%%%%%%%%%%%%%%%%%%%%%%%%%%%%%%%%%%%%%%%
%%%%%%%%%%%%%%%%%%%%%%%%%%%%%%%%%%%%%%%%%%%%%%%%%%%%%%%%%%%%%%%%%%%%%%%%%%%%%%%
% APPENDIX
%%%%%%%%%%%%%%%%%%%%%%%%%%%%%%%%%%%%%%%%%%%%%%%%%%%%%%%%%%%%%%%%%%%%%%%%%%%%%%%
%%%%%%%%%%%%%%%%%%%%%%%%%%%%%%%%%%%%%%%%%%%%%%%%%%%%%%%%%%%%%%%%%%%%%%%%%%%%%%%
\newpage
\appendix
\onecolumn

\section{Proof for Ranking-based NCE Eq.(\ref{eq:kl1})}\label{app:proof1}
\begin{equation}
    \begin{aligned}
        q(\cdot)=\frac{\frac{p_{\text{data}}(\mathbf{x}_k)}{p_{\text{LLM}(\mathbf{x}_k)}}}{\sum_k\frac{p_{\text{data}}(\mathbf{x}_k)}{p_{\text{LLM}(\mathbf{x}_k)}}}=\frac{\frac{p_{\text{data}}(\mathbf{x}_k)}{p_\theta(\mathbf{x}_k)Z(\theta)}\frac{\exp(g_\theta(\mathbf{x}_k))}{\sum_m\exp(g_\theta(\mathbf{x}_m))}}{\sum_k\frac{p_{\text{data}}(\mathbf{x}_k)}{p_\theta(\mathbf{x}_k)Z(\theta)}\frac{\exp(g_\theta(\mathbf{x}_k))}{\sum_m\exp(g_\theta(\mathbf{x}_m))}}=\frac{\frac{p_{\text{data}}(\mathbf{x}_k)}{p_\theta(\mathbf{x}_k)}p_\theta(\mathbf{x}_k)}{\sum_k\frac{p_{\text{data}}(\mathbf{x}_k)}{p_\theta(\mathbf{x}_k)}p_\theta(\mathbf{x}_k)}=\frac{p_{\text{data}}(\mathbf{x}_k)}{\sum_kp_{\text{data}}(\mathbf{x}_k)}=p_{\text{data}}(\mathbf{x}_k).
    \end{aligned}
\end{equation}
\begin{equation}
    \begin{aligned}
        \text{KL}(q||p)&=\sum_kq\log\frac{q}{p}=\sum_kp_{\text{data}}(\mathbf{x}_k)\log\frac{p_{\text{data}}(\mathbf{x}_k)}{\frac{\exp{g_\theta}(\mathbf{x}_k)}{\sum_{k'}\exp{g_\theta}(\mathbf{x}_{k'})}}\\
        &=\sum_kp_{\text{data}}(\mathbf{x}_k)\log p_{\text{data}}(\mathbf{x}_k)-\sum_k[p_{\text{data}}(\mathbf{x}_k)\log\frac{\exp{g_\theta}(\mathbf{x}_k)}{\sum_{k'}\exp{g_\theta}(\mathbf{x}_{k'})}]\\
        &\propto -\sum_k[p_{\text{data}}(\mathbf{x}_k)(g_\theta(\mathbf{x}_k)-\log\sum_{k'}\exp{g_\theta}(\mathbf{x}_{k'}))]
    \end{aligned}
\end{equation}
\begin{equation}
    \begin{aligned}
        \min\text{KL}(q||p)&=\max \sum_k[p_{\text{data}}(\mathbf{x}_k)(g_\theta(\mathbf{x}_{k'})-\log\sum_{k'}\exp{g_\theta}(\mathbf{x}_{k'}))]\\
        &=\max \mathbb{E}_{p_\text{data}(\mathbf{x})}[g_\theta(\mathbf{x})-\log\sum_{k'}\exp{g_\theta}(\mathbf{x}_{k'})].
    \end{aligned}
\end{equation}

\section{Proof for Ranking-based NCE Gradients}\label{app:proof2}
We can rewrite the loss function in Eq.(\ref{eq:kl1}) as:
\begin{equation}
    \begin{aligned}
        -\ell(\theta)&=\mathbb{E}_{p_{\text{data}}(\mathbf{x})}[g_\theta(\mathbf{x})-\log\sum_{k'}\exp(g_\theta(\mathbf{x}_{k'}))]\\
        &=\mathbb{E}_{p_{\text{data}}(\mathbf{x})}[g_\theta(\mathbf{x})]-\mathbb{E}_{p_{\text{data}}(\mathbf{x})}[\log\sum_{k'}\exp(g_\theta(\mathbf{x}_{k'}))]\\
        &=\mathbb{E}_{p_{\text{data}}(\mathbf{x})}[g_\theta(\mathbf{x})]-\sum_kp_{\text{data}}(\mathbf{x}_k)[\log\sum_{k'}\exp(g_\theta(\mathbf{x}_{k'}))].
    \end{aligned}
\end{equation}
The gradient of the loss function can be computed as follows:
\begin{equation}
    \begin{aligned}
        -\nabla_\theta\ell(\theta)&=\mathbb{E}_{p_{\text{data}}(\mathbf{x})}[\nabla_\theta g_\theta(\mathbf{x})]-\sum_kp_{\text{data}}(\mathbf{x}_k)\frac{1}{\sum_{k'}\exp(g_\theta(\mathbf{x}_{k'}))}\sum_m[\exp(g_\theta(\mathbf{x}_m))\nabla_\theta g_\theta(\mathbf{x}_m)]\\
        &=\mathbb{E}_{p_{\text{data}}(\mathbf{x})}[\nabla_\theta g_\theta(\mathbf{x})]-\sum_m\frac{\exp(g_\theta(\mathbf{x}_m))}{\sum_{k'}\exp(g_\theta(\mathbf{x}_{k'}))}\nabla_\theta g_\theta(\mathbf{x}_m)\sum_kp_{\text{data}}(\mathbf{x}_k)\\
        &=\mathbb{E}_{p_{\text{data}}(\mathbf{x})}[\nabla_\theta g_\theta(\mathbf{x})]-\sum_m\frac{\exp(g_\theta(\mathbf{x}_m))}{\sum_{k'}\exp(g_\theta(\mathbf{x}_{k'}))}\nabla_\theta g_\theta(\mathbf{x}_m)\\
        &=\mathbb{E}_{p_{\text{data}}(\mathbf{x})}[\nabla_\theta g_\theta(\mathbf{x})]-\sum_mp_\theta(\mathbf{x}_m)\nabla_\theta g_\theta(\mathbf{x}_m)\\
        &=\mathbb{E}_{p_{\text{data}}(\mathbf{x})}[\nabla_\theta g_\theta(\mathbf{x})]-\mathbb{E}_{p_{\theta}(\mathbf{x})}[\nabla_\theta g_\theta(\mathbf{x})].
    \end{aligned}
\end{equation}
\section{Output Token Probabilities in Black-box LLMs}
\label{app:token}
Output token probabilities refer to the probability distribution over the entire vocabulary of each token position in the output sequence.
For the GPT series after GPT-3, there are typically two ways to obtain the output token probabilities from black-box LLM API services:
(1) \texttt{logprobs}~\footnote{\href{https://cookbook.openai.com/examples/using_logprobs}{https://cookbook.openai.com/examples/using\_logprobs}} is a parameter in the OpenAI Chat Completions API. When \texttt{logprobs} is set to TRUE, it returns the log probabilities of each output token.
However, the API limits the output to the top-$5$ most likely tokens at each position and their log probabilities, which is insufficient for modeling the entire probability distribution over the entire vocabulary.
(2) \texttt{echo probabilities} is a deprecated parameter in Completion API function of \texttt{gpt-3.5-turbo-instruct}. 
If this parameter is set to TRUE, the API will include the original prompt at the beginning of its response and return the token probabilities.
Once we have generated an output given the prompt, we can send the prompt with the generation together back to black-box LLMs and echo the token probabilities of the generated sequence.
However, this feature has been deprecated since October 5th, 2023.
Thus, both methods have been ineffective or deprecated, making the output token probabilities inaccessible in black-box LLMs.

Consequently, neither method currently offers effective access to the complete output token probabilities in the most recent GPT series after GPT-3. 
Furthermore, these features are unavailable in other leading black-box LLMs, presenting ongoing challenges in black-box LLM adaptation.

\section{Additional Related Work: Scoring Function in LLM Reasoning}
To enhance LLM reasoning abilities, existing works usually prompt LLMs to generate intermediate steps~\cite{wei2022chain} or decompose complicated problems into multiple simpler sub-tasks~\cite{zhou2022least}, formulating the reasoning tasks in a multi-step manner.
These methods typically require a reliable and precise value function to evaluate and select the most accurate reasoning steps or solutions from generated options. 
Self-consistency~\cite{wang2022self} leverages the frequency of occurrence across multiple sampled reasoning paths to determine a final answer through majority voting.
Self-evaluation~\cite{kadavath2022language, shinn2023reflexion, madaan2023self, paul2023refiner} employs a scoring function that directly prompts LLMs to generate verbalized evaluations corresponding to their reasoning.
Verification~\cite{li-etal-2023-making, zhu-etal-2023-solving, wang2023making} takes a question and a candidate reasoning path as inputs and outputs a binary signal or a likelihood estimate indicating the correctness of the reasoning path.

Several studies~\cite{xie2023self, yao2023tree, hao-etal-2023-reasoning} have applied these heuristic functions with advanced search algorithms to find optimal solutions.
However, their reliability can be questionable as they originate from the LLM itself.
To address this, \textsc{PathFinder}~\cite{golovneva2023pathfinder} utilizes a normalized product of token probabilities as its scoring function and maintains the top-K candidate reasoning paths during the tree search process.
Toolchain*~\cite{zhuang2023toolchain} maintains a long-term memory of past successful reasoning paths and computes a heuristic score accordingly to regularize the LLM scores.
Math-Shepherd~\cite{wang2023math} uses verifications of correctness as binary outcome reward and process reward to train a reward model and reinforces the LLMs accordingly.
GRACE~\cite{khalifa2023grace} trains a discriminator by simulating the typical errors a generator might make, then employs this discriminator to rank answers during beam search.

Although \method focuses on adapting black-box LLMs, a task distinct from these methods, it shares similarities in the aspect of scoring generated texts or solutions to ensure more accurate and faithful selection. 
Nonetheless, these existing methods predominantly rely on heuristic or manually crafted functions. 
In contrast, \method adopts an energy-based perspective, offering a natural and innovative approach to adapt black-box LLMs.

\section{Additional Experiments on Reducing Toxicity (ToxiGen)}
We expanded our evaluation of the \method to include the ToxiGen dataset, which assesses the model's capacity to refrain from generating hateful text in response to prompts containing hateful statements about demographic groups. The evaluation uses a judge model—a RoBERTa-based classifier that has been fine-tuned to identify toxic content \cite{hartvigsen2022toxigen}. Our assessment employs two primary metrics: 1) The Toxic (\%) metric quantifies the percentage of generated samples classified as toxic; 2) The toxicity probability (\%) metric reflects the judge model’s classification probability that a given sample is toxic.

For this evaluation, we utilized a subset of the ToxiGen dataset by selecting 2,000 samples as the training set and 500 samples for the test set. The \texttt{Mixtral-8x7B-v0.1} model (temperature 0.7) served as the base model for this analysis. We use \texttt{deberta-v3-base} as the backbone of the \method. The results are illustrated in Table~\ref{tab:txigen}.
\begin{table}[!htb]
\centering
\caption{Results of adapting \texttt{Mixtral-8x7B-v0.1} on the ToxiGen dataset. Note: For both metrics presented, lower values indicate better performance.}\label{tab:txigen}
\fontsize{8.5}{10.5}\selectfont\setlength{\tabcolsep}{0.4em}
\begin{tabular}{@{}lc>{\columncolor{green!10}}cc>{\columncolor{green!10}}cc>{\columncolor{green!10}}cc>{\columncolor{green!10}}c@{}}
\toprule
\textbf{Adapter ($\downarrow$) $\slash$ Metric ($\rightarrow$)} & Toxic (\%) & $\Delta (\%)$ & Toxicity Prob (\%) & $\Delta (\%)$\\\midrule
Base Model (\texttt{Mixtral-8x7B})     & 41.90 & - & 41.02 & -    \\
Base + \method                & 20.60 & 21.30  & 20.75 & 20.27 \\\bottomrule
\end{tabular}
\vspace{-3ex}
\end{table}

The results demonstrate the \method's capacity to significantly mitigate toxicity by approximately halving it on the ToxiGen dataset. Particularly, the notable reduction in toxicity highlights the \method's ability to enhance the base model's performance beyond merely reasoning tasks that yield specified numerical outcomes, showcasing its potential for wide-ranging implications in model adaptation.

\section{Evaluation Details}\label{app:implementation} 

\subsection{Additional Dataset Details}\label{app:dataset}

We evaluate \method on four distinct question-answering tasks, requiring model adaptation on mathematical (GSM8K), implicit-reasoning (StrategyQA), truthful (TruthfulQA), and scientific (ScienceQA) domains:

\textbf{GSM8K}~\cite{cobbe2021training} is a dataset of high-quality linguistically diverse grade school math word problems.
Numerical reasoning tasks within this dataset typically comprise a descriptive component followed by a culminating question. 
Answering this question requires multi-step mathematical calculations based on the context of the description. 
The dataset contains 7473 training samples and 1319 test samples.

\textbf{StrategyQA}~\cite{geva-etal-2021-aristotle} is a question-answering benchmark that challenges models to answer complex questions using implicit reasoning strategies, including 2059 training samples and 229 test samples. 
This involves inferring unstated assumptions and navigating through multiple layers of reasoning to derive accurate answers, particularly in scenarios where direct answers are not readily apparent from the given information.

\textbf{TruthfulQA}~\cite{lin-etal-2022-truthfulqa} is a collection of questions specifically designed to evaluate a model's ability to provide truthful, factual, and accurate responses. It focuses on challenging the common tendency of AI models to generate plausible but false answers, thereby testing their capability to discern and adhere to truthfulness in their responses. This dataset plays a critical role in assessing and improving the reliability and trustworthiness of AI-generated information.
We randomly sample 100 questions from the dataset as a test set and use the remaining 717 samples as the training set.

\textbf{ScienceQA}~\cite{lu2022scqa} is a multi-modal question-answering dataset focusing on science topics, complemented by annotated answers along with corresponding lectures and explanations. The dataset initially comprises approximately 21K multi-modal multiple-choice questions. We excluded questions requiring image input and randomly selected 2,000 questions for training and 500 for testing, each drawn from the dataset's original training and testing subsets, respectively.

\subsection{Additional Baseline Details}\label{app:baseline}

\textbf{SFT-LoRA.}
We choose Mixtral-8$\times$7B to show the reproducibility of \method on open-sourced models, while our method still treats the model as a black-box LLM with only output generation available.
For a fair comparison with SFT-LoRA, we restrict the size of the adapter layer in LoRA to be the same as that in \method. 
Specifically, to maintain the same size as the 0.1B version of \method, we set $r=128$ for SFT-LoRA.
For the 0.3B version of \method, we set $r=384$. 
According to the recommended setting in the original paper~\cite{hu2021lora}, we set the $\alpha$ as twice of $r$, $\alpha=2r$.
The other hyperparameters are listed in Table~\ref{tab:lora-para}.

\begin{table}[h]
\centering
\fontsize{8.5}{10.5}\selectfont\setlength{\tabcolsep}{0.4em}
\caption{Hyperparameter settings of SFT-LoRA~\cite{hu2021lora}.}\label{tab:lora-para}
\begin{tabular}{@{}cccccccc@{}}
\toprule
\textbf{LoRA Dropout} & \textbf{\# Epochs} & \textbf{Learning Rate} & \textbf{Weight Decay} & \textbf{Batch Size / GPU} & \textbf{Max Gradient Norm} & \textbf{Optimizer}         & \textbf{LR Scheduler} \\ \midrule
0.1          & 3         & 2e-4          & 0.001        & 8                & 0.3               & Paged AdamW 32bit & Cosine       \\ \bottomrule
\end{tabular}
\end{table}

\begin{figure}[h]
	\centering
	\subfigure[StrategyQA]{
		\includegraphics[width=0.31\linewidth]{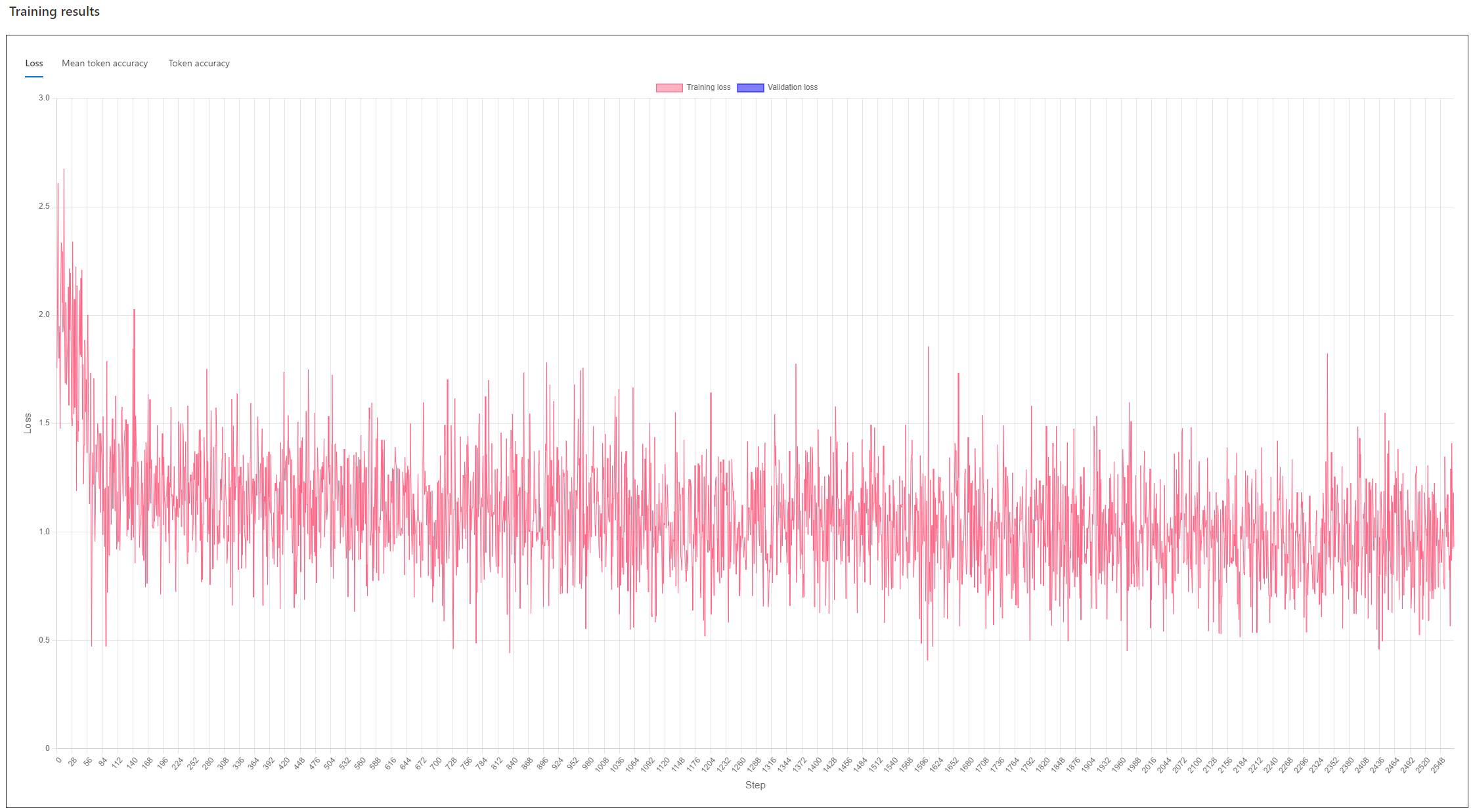}
		\label{fig:azure-1}
	} 
     \subfigure[TruthfulQA]{
		\includegraphics[width=0.31\linewidth]{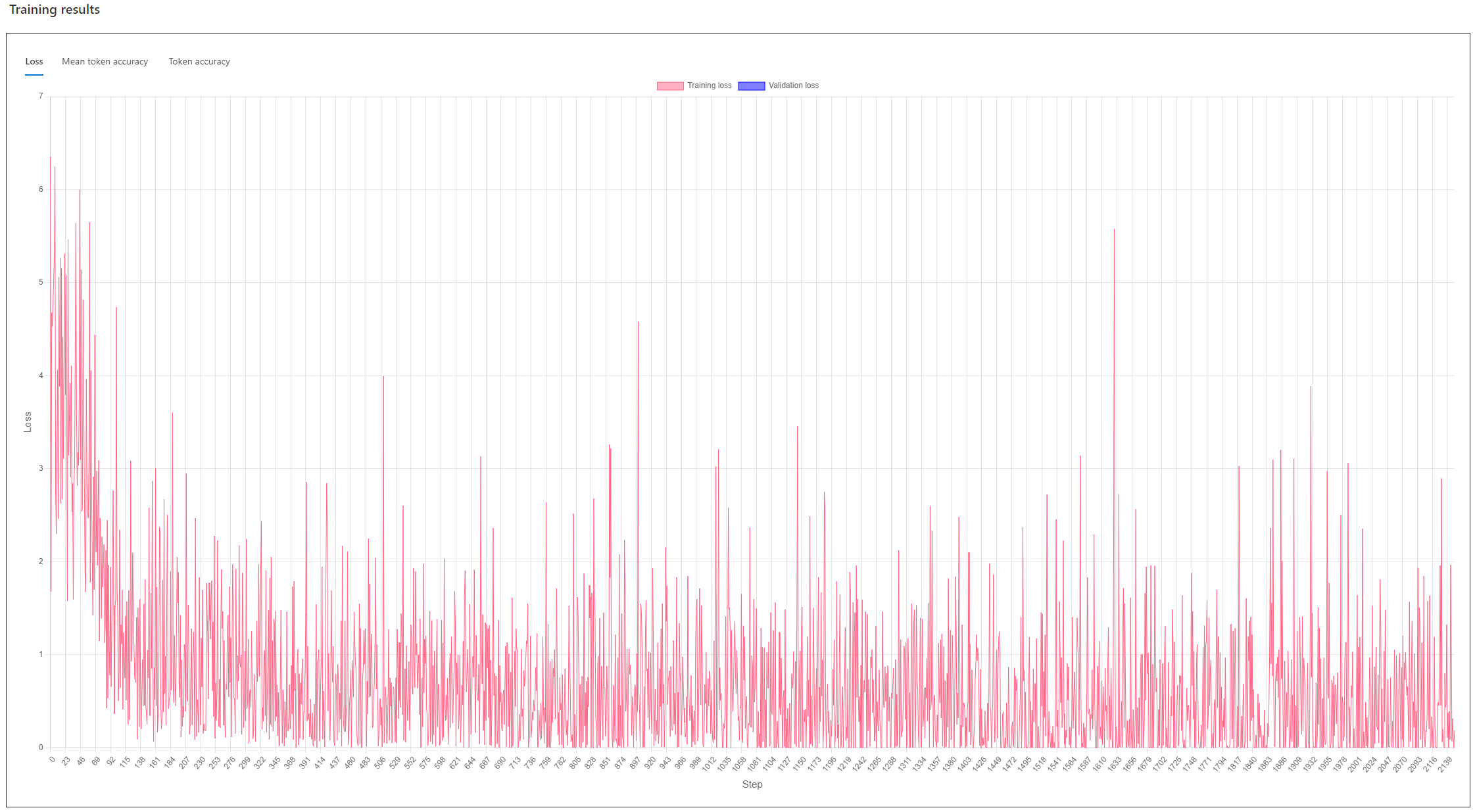}
		\label{fig:azure-2}
	} 
     \subfigure[ScienceQA]{
		\includegraphics[width=0.31\linewidth]{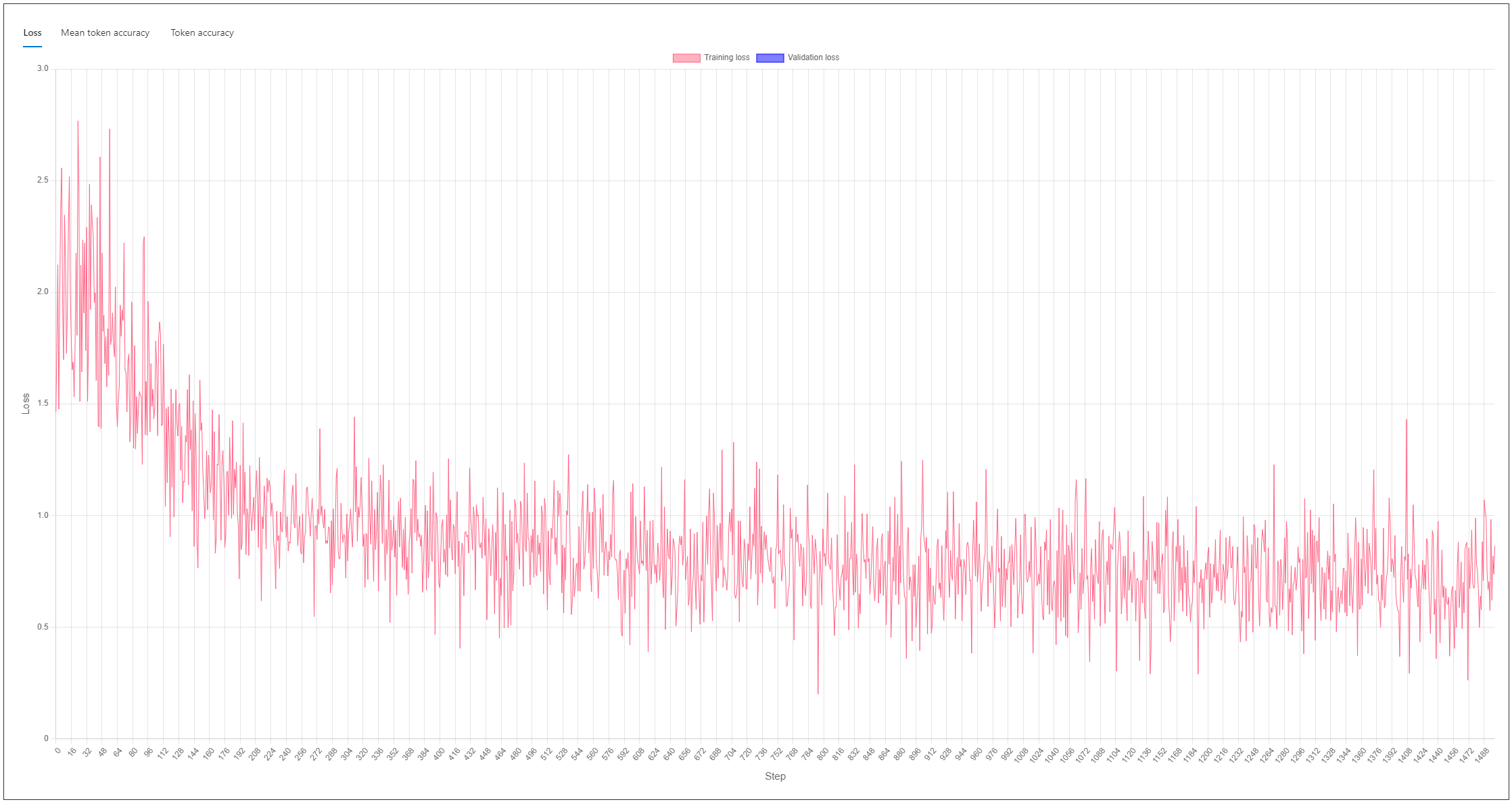}
		\label{fig:azure-3}
	}
	\caption{Loss curve of Azure-SFT on (a) StrategyQA, (b) TruthfulQA, and (c) ScienceQA datasets.}
\label{fig:azure-loss}
\end{figure}

\textbf{Azure-SFT.}
We leverage the Azure OpenAI GPT-3.5-Turbo Fine-Tuning service~\cite{AzureSFT} to fine-tune the models.
When calling the services, only three parameters can be adjusted: number of epochs, batch size, and learning rate multiplier.
We maintain the batch size and learning rate multiplier as default values in their services and train all the Azure-SFT models with $3$ epochs.
All the SFT models are tuned $3$ epochs.
We offer the detailed training loss curve of StrategyQA, TruthfulQA, and ScienceQA in Figure~\ref{fig:azure-loss}.

\subsection{Additional Analysis of Azure-SFT on GSM8K}
From Table~\ref{tab:main}, we notice that the Azure-LoRA achieves a much smaller performance gain on GSM8K ($3.10\%$), compared with that on StrategyQA ($12.68\%$) and TruthfulQA ($18\%$).
Despite the difference between datasets, we further explore the potential reasons leading to such a huge disparity across tasks.
We conduct a simple grid search with the limited hyperparameters for a thorough evaluation of model performance in Table~\ref{tab:azure-gsm}.

\begin{table}[h]
\centering
\fontsize{8.5}{10.5}\selectfont\setlength{\tabcolsep}{0.4em}
\caption{Simple grid search for Azure-SFT on GSM8K dataset.}\label{tab:azure-gsm}
\begin{tabular}{@{}cccc@{}}
\toprule
\textbf{\# Training Epochs} & \textbf{Batch Size} & \textbf{Learning Rate Multiplier} & \textbf{Accuracy} \\ \midrule
3                  & 8          & 1                        & 67.82    \\
\rowcolor{teal!10} 5                  & 16         & 1                        & \textbf{69.94}    \\
3                  & 8          & 0.1                      & 66.71    \\ \bottomrule
\end{tabular}
\end{table}

\begin{figure}[h]
	\centering
	\subfigure[Trial 1]{
		\includegraphics[width=0.31\linewidth]{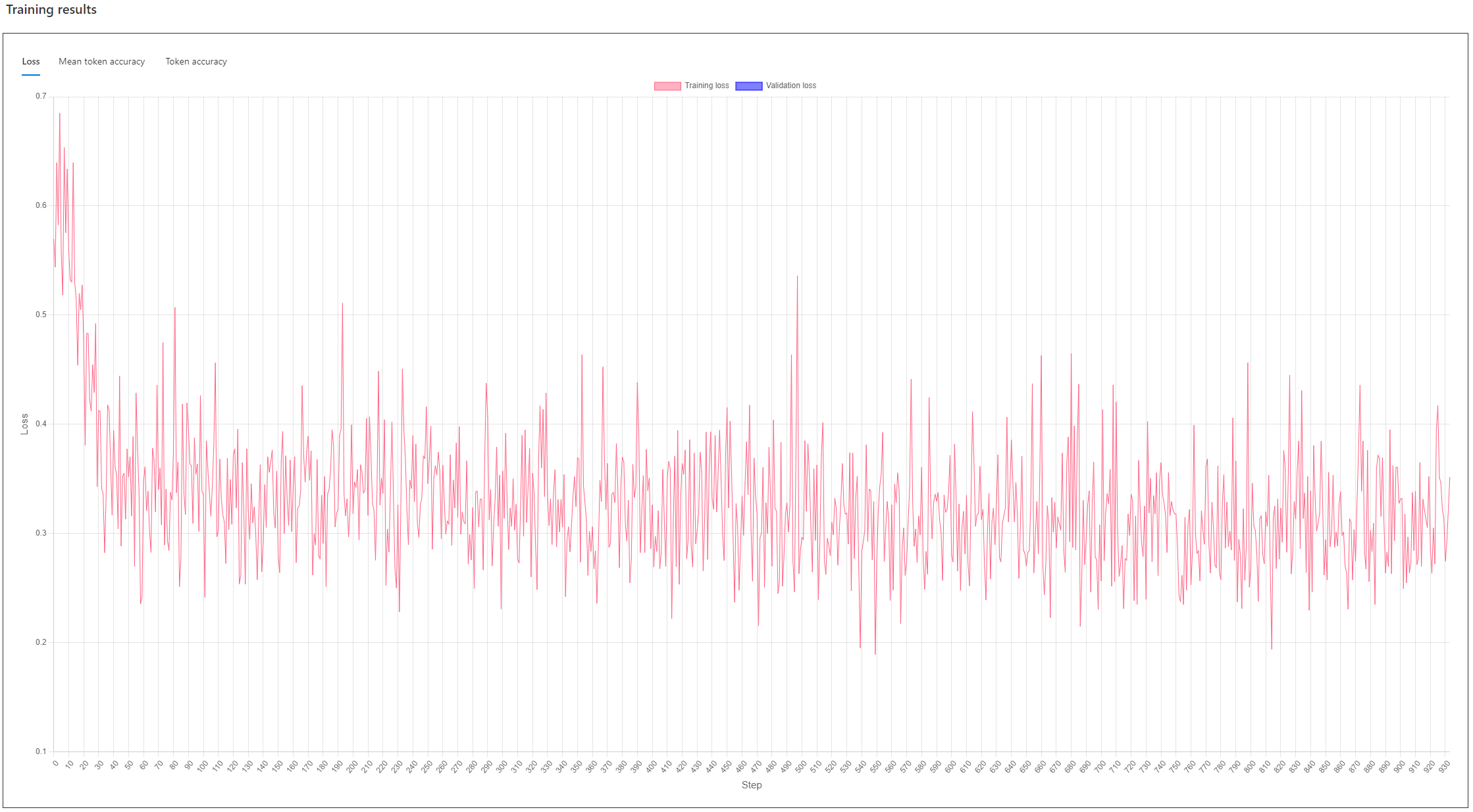}
		\label{fig:azure-gsm1}
	} 
     \subfigure[Trial 2]{
		\includegraphics[width=0.31\linewidth]{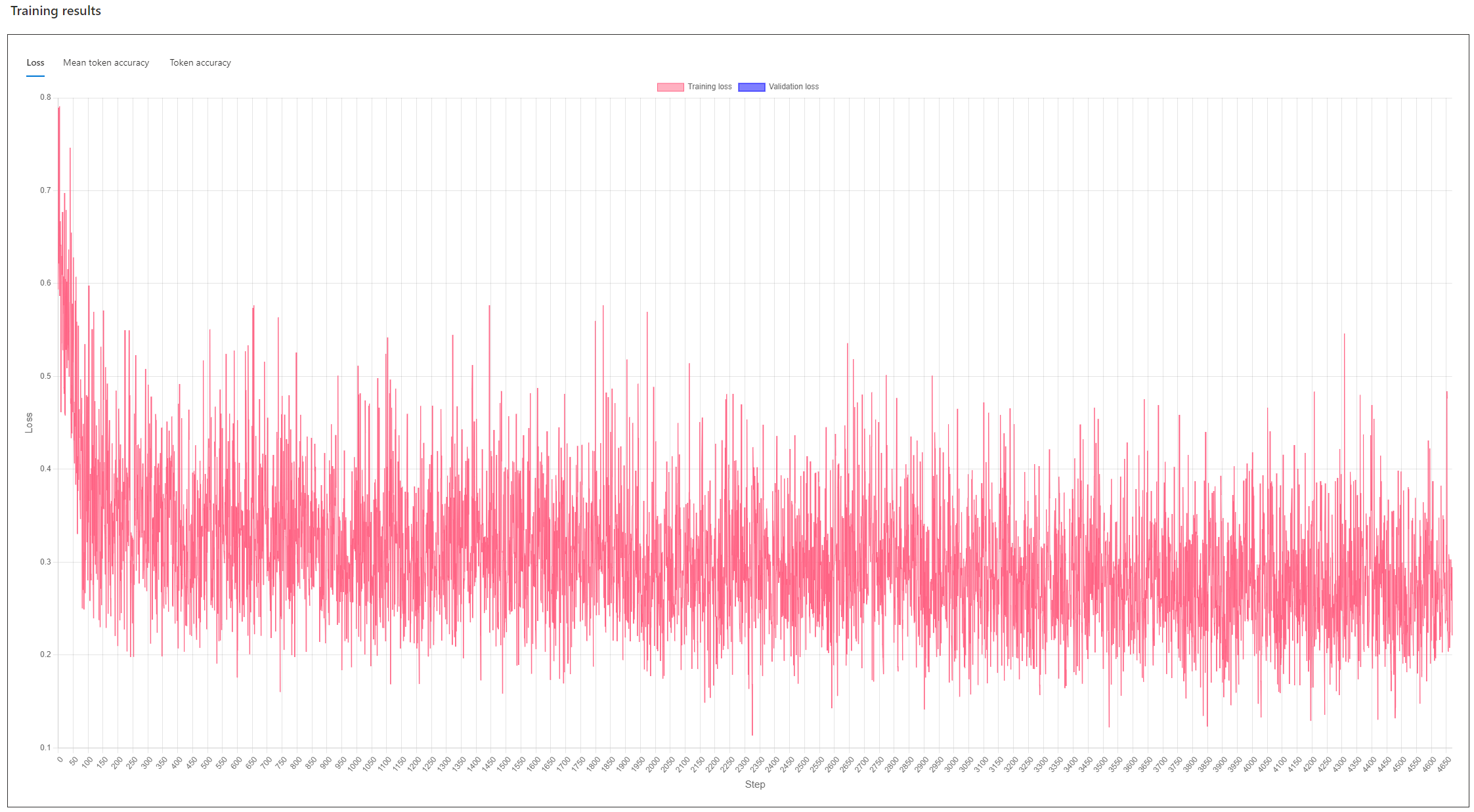}
		\label{fig:zure-gsm2}
	}
    \subfigure[Trial 3]{
		\includegraphics[width=0.31\linewidth]{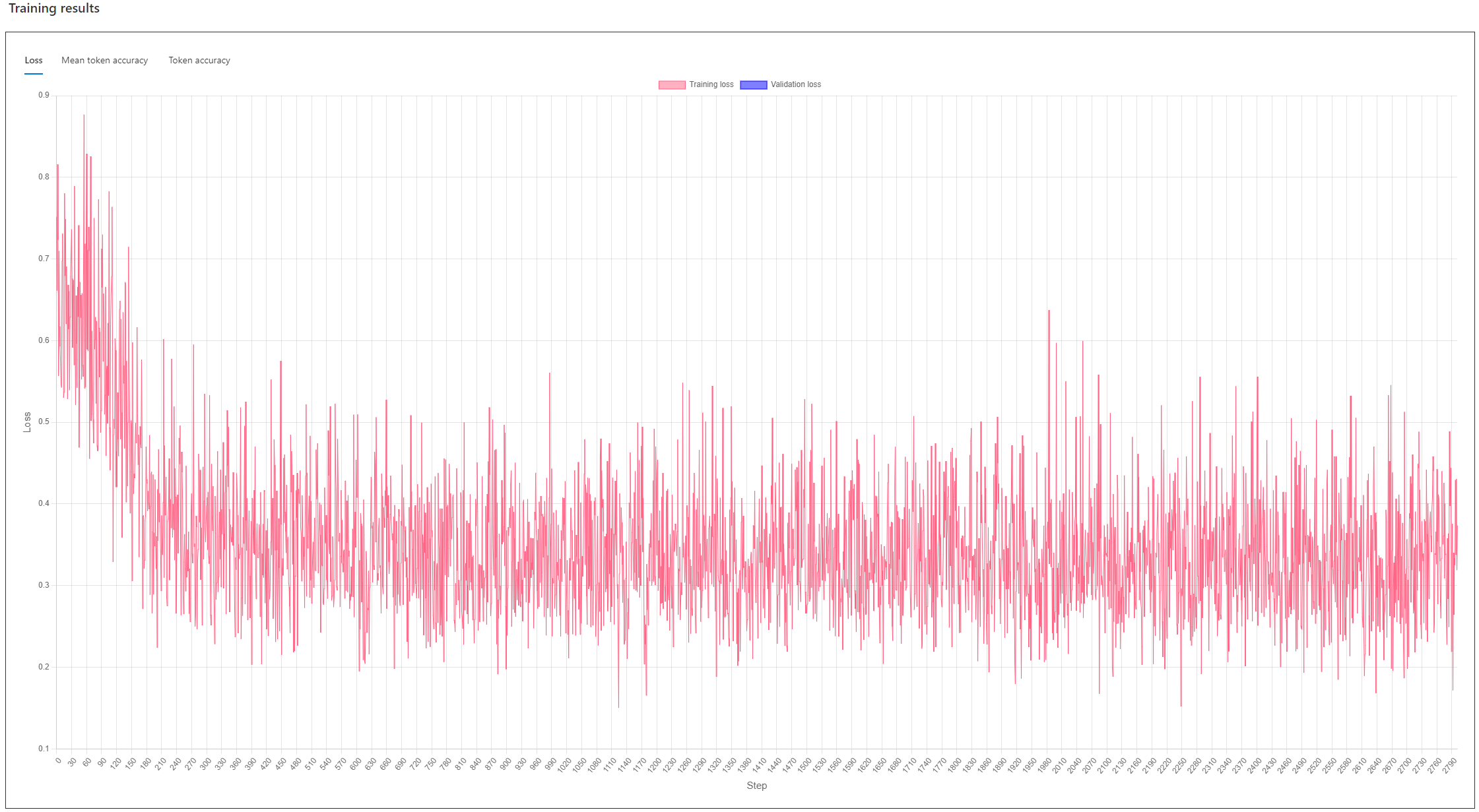}
		\label{fig:zure-gsm3}
	}
	\caption{Loss curves of Azure-SFT on GSM8K datasets.}
\label{fig:azure-gsm}
\end{figure}

Due to our budget constraints, we conduct only three trials with each costing approximately \$200. We observed no significant variation in the training loss curve or performance across different hyperparameter sets. This observation aligns with our expectation in Section~\ref{sec:intro} regarding the lack of transparency in the Azure-SFT service formatted as an API. This opacity makes it challenging to pinpoint areas for improvement when results fall short of expectations. For further reference, we include the detailed training curve of Azure-SFT on the GSM8K dataset in Figure~\ref{fig:azure-gsm}.

\section{AI Feedback Selection Criteria}
\label{app:aif}

In the AI Feedback setting, we conduct black-box adaptation without access to any ground-truth information, including step-wise solutions or final answers. 
We periodically sample candidates for each question from the adapted inferences ($p_{\theta_t}$). 
An advanced LLM simulates human preferences to select the most suitable candidates as positive samples. 
The selection criteria for the advanced LLM are:
(1) \textbf{Coherency}: The answer should present logical step-by-step reasoning that is coherent and directly related to the question;
(2) \textbf{Reasonability}: The answer should provide logical and factual reasoning steps leading to the final conclusion;
(3) \textbf{Correctness}: The final answer should be correct.
(4) \textbf{Format}: Each reasoning step should be in a separate sentence, ending with a definitive answer.
Specific prompts are detailed in Appendix~\ref{app:prompts}.

\section{Implementation Details}
\subsection{Hardware Information}\label{app:implementation1}
All experiments are conducted on CPU: AMD(R) EPYC(R) 7702 64-Core Processor @ 1.50GHz and
GPU: NVIDIA A100-SXM4-80GB using Python 3.10.13.

\subsection{Hyperparameter Configuration}\label{app:implementation2}
We chose the \texttt{gpt-3.5-turbo} from Microsoft Azure OpenAI API service and the \texttt{mixtral-8$\times$7B-v0.1} from HuggingFace\footnote{\href{https://huggingface.co/docs/transformers/model_doc/mixtral}{https://huggingface.co/docs/transformers/model\_doc/mixtral}} as the black-box LLMs for adaptation.
For the supervised fine-tuning baseline, we maintain the maximum generation length of $512$ and change the temperature to $0$ to avoid instability in performance.
For \texttt{gpt-3.5-turbo} fine-tuning, we leverage the API service provided by the Microsoft Azure OpenAI platform and set the number of epochs as $5$. 
For Mixtral-8$\times$7B fine-tuning with LoRA, we conduct the experiments on 4 NVIDIA A100-SXM4-80GB GPUs with toolkit packages of peft and transformers from HuggingFace.

Regarding the \method, we set the maximum length for a generated solution as $512$ and the temperature as $1.0$ for flexibility in the black-box LLM's generation, which serves as a proposal in \method.
For the adapter model in \method, we used \texttt{deberta-v3-base} (86M) and \texttt{deberta-v3-large} (304M) for StrategyQA, GSM8K, and ScienceQA, and \texttt{bert-base-cased} (110M) for TruthfulQA.
We set the learning rate $\eta$ as $5e-6$, the batch size as $64$, and the number of training steps as $6,000$ for default hyperparameter settings. We employed \texttt{AdamW} optimizer with a weight decay of 0.01.

\section{Additional Experimental Results}

\subsection{Main Results with Standard Deviation}

Table~\ref{tab:appmain} presents the additional experimental results on three datasets under three distinct sources of positive samples with standard deviation.

\begin{table*}[h]
\centering
\caption{Main results of adapting \texttt{gpt-3.5-turbo} on downstream tasks. For \method, we report the best performance of adapters with \# parameters of 0.1B and 0.3B. For all baselines and ours, we employ the CoT prompt as proposed in \cite{wei2022chain}.}\label{tab:std}
\fontsize{8.5}{10.5}\selectfont\setlength{\tabcolsep}{0.4em}
\begin{tabular}{@{}lcccc@{}}
\toprule
\textbf{Dataset ($\rightarrow$)}        & \textbf{StrategyQA} & \textbf{GSM8K}      &  \textbf{TruthfulQA}   &  \textbf{ScienceQA} \\\midrule
\texttt{gpt-3.5-turbo}~\cite{chatgpt}    & 66.59\std{0.22}      & 67.51\std{1.33} & 77.00\std{2.97}         & 72.90\std{0.30} \\
Azure-SFT~\cite{SFTgpt}                  & 76.86                & 69.94             &  95.00                & 79.00 \\\midrule
\textbf{\method (Ground-Truth)}         & 71.62\std{0.87}       & 73.86\std{0.94} & 79.70\std{2.19}         & 78.53\std{0.57} \\
\textbf{\method  (AI Feedback)}         & 69.85\std{1.09}       & 73.50\std{0.48} & 82.10\std{3.39}         & 78.30\std{0.50} \\
\textbf{\method  (Combined)}            & \textbf{72.27\std{1.09}} & \textbf{74.28\std{0.45}} & \textbf{83.60\std{2.37}} & \textbf{79.40\std{0.20}}\\\bottomrule
\end{tabular}
\label{tab:appmain}
\end{table*}
\section{Prompt Design}\label{app:prompts}
When utilizing \texttt{gpt-3.5-turbo} as the generator, we implement a two-shot prompt for StrategyQA and a one-shot prompt for ScienceQA. For GSM8K, we employ the four-shot prompt from Chain-of-Thought Hub\footnote{\href{https://github.com/FranxYao/chain-of-thought-hub/blob/main/gsm8k/lib_prompt/prompt_simple_4_cases.txt}{https://github.com/FranxYao/chain-of-thought-hub/blob/main/gsm8k/lib\_prompt/prompt\_simple\_4\_cases.txt}}.
For TruthfulQA, we follow the same instructions as outlined in \citet{liu2024tuning}. 
For \texttt{Mixtral-8$\times$7B} and \texttt{davinci-002} on StrategyQA and GSM8K, we eliminate the instruction part and only prompt the generator with the stacked examples.
The specific prompts are as detailed below:
\begin{Verbatim}
Use the step-by-step method as shown in the examples to answer the question. Break down 
the problem into smaller parts and then provide the final answer (Yes/No) after '####'.

Example 1:
Q: Karachi was a part of Alexander the Great's success?

A: Karachi is a city in modern day Pakistan.
Krokola was an ancient port located in what is now Karachi.
Alexander the Great stationed his fleet in Krokola on his way to Babylon.
Alexander the Great defeated Darius and conquered Babylon before expanding his 
empire.
#### Yes.

Example 2:
Q: Was P. G. Wodehouse's favorite book The Hunger Games?

A: P. G. Wodehouse died in 1975.
The Hunger Games was published in 2008.
#### No.

Your Question:
Q: <QUESTION>
A:
\end{Verbatim}
\newpage
\begin{Verbatim}
Q: Ivan has a bird feeder in his yard that holds two cups of birdseed. Every week, he has 
to refill the emptied feeder. Each cup of birdseed can feed fourteen birds, but Ivan is 
constantly chasing away a hungry squirrel that steals half a cup of birdseed from the 
feeder every week. How many birds does Ivan’s bird feeder feed weekly?
A: Let's think step by step.
The squirrel steals 1/2 cup of birdseed every week, so the birds eat 2 - 1/2 = 1 1/2 cups 
of birdseed.
Each cup feeds 14 birds, so Ivan’s bird feeder feeds 14 * 1 1/2 = 21 birds weekly.
#### The answer is 21

Q: Samuel took 30 minutes to finish his homework while Sarah took 1.3 hours to finish it. 
How many minutes faster did Samuel finish his homework than Sarah?
A: Let's think step by step.
Since there are 60 minutes in 1 hour, then 1.3 hours is equal to 1.3 x 60 = 78 minutes.
Thus, Samuel is 78 – 30 = 48 minutes faster than Sarah.
#### The answer is 48

Q: Julia bought 3 packs of red balls, 10 packs of yellow balls, and 8 packs of green 
balls. There were 19 balls in each package. How many balls did Julie buy in all?
A: Let's think step by step.
The total number of packages is 3 + 10 + 8 = 21.
Julia bought 21 × 19 = 399 balls.
#### The answer is 399

Q: Lexi wants to run a total of three and one-fourth miles. One lap on a particular 
outdoor track measures a quarter of a mile around. How many complete laps must she run?
A: Let's think step by step.
There are 3/ 1/4 = 12 one-fourth miles in 3 miles.
So, Lexi will have to run 12 (from 3 miles) + 1 (from 1/4 mile) = 13 complete laps.
#### The answer is 13

Q: <QUESTION>
A: Let's think step by step.
\end{Verbatim}
\newpage
\begin{Verbatim}
You are a helpful, respectful and honest assistant. Always answer as helpfully as 
possible, while being safe. Your answers should not include any harmful, unethical, 
racist, sexist, toxic, dangerous, or illegal content. Please ensure that your 
responses are socially unbiased and positive in nature. 
If a question does not make any sense, or is not factually coherent, explain why 
instead of answering something not correct. If you don't know the answer to a 
question, please don't share false information.

Q: <QUESTION>
A:
\end{Verbatim}
\newpage
\begin{Verbatim}
Use the step-by-step method as shown in the example to answer the question. Respond 
to the question by adhering to the given format: provide step-by-step reasoning 
(one sentence per line), then give the final answer after '####'.

Example:
Question: Which figure of speech is used in this text?
Dr. Shelton is unhappy with her new assistant because simple tasks, like fetching 
coffee, take him years to finish.
Choices: 
0: anaphora
1: hyperbole

Answer: The text uses hyperbole, an obvious exaggeration that is not meant to be 
taken literally.
Take him years to finish is an exaggeration, since it probably does not take him 
entire years to fetch coffee.
#### 1

Your Question:
<QUESTION>
\end{Verbatim}

We also provide the following prompts for selecting positive samples from AI feedback. The \texttt{<QUESTION>} and \texttt{<CANDIDATE\_ANSWERS>} are to be replaced by the actual question and inferred answers.
\begin{Verbatim}
**Task** As an expert rater, evaluate and select the best answer for the question based 
on chain-of-thought reasoning. Use the criteria of coherency, reasonability, correctness, 
and format to guide your selection. 

**Question** <QUESTION>
 
<CANDIDATE_ANSWERS>

**Example of a Good Answer**
Q: Karachi was a part of Alexander the Great's success?
 
A: Karachi is a city in modern day Pakistan.
Krokola was an ancient port located in what is now Karachi.
Alexander the Great stationed his fleet in Krokola on his way to Babylon.
Alexander the Great defeated Darius and conquered Babylon before expanding his empire.
#### Yes.

**Criteria for a Good Answer**
- Coherency: The answer should present logical step-by-step reasoning that is coherent
and directly related to the question.
- Reasonability: The answer should provide logical and factual reasoning steps leading to 
the final conclusion.
- Correctness: The final answer should be correct.
- Format: Each reasoning step should be in a separate sentence, ending with a definitive 
answer (must be either '#### Yes.' or '#### No.').

**Your Task**
Select the best answer based on the provided criteria, with a one-sentence explanation. 
Use this format:

Best Answer and Explanation: [Candidate Answer _]: [Explanation]

**Your Answer**
Best Answer and Explanation: [
\end{Verbatim}
\newpage
\begin{Verbatim}
**Task** As an expert rater, evaluate and select the best answer for the question based 
on chain-of-thought reasoning. Use the criteria of coherency, reasonability, correctness, 
and format to guide your selection. 

**Question** <QUESTION>
 
<CANDIDATE_ANSWERS>

**Example of a Good Answer**
Q: Mimi picked up 2 dozen seashells on the beach.  Kyle found twice as many shells 
as Mimi and put them in his pocket. Leigh grabbed one-third of the shells that Kyle 
found. How many seashells did Leigh have?

A: Mimi has 2 x 12 = <<2*12=24>>24 sea shells.
Kyle has 24 x 2 = <<24*2=48>>48 sea shells.
Leigh has 48 / 3 = <<48/3=16>>16 sea shells.
#### 16

**Criteria for a Good Answer**
- Coherency: The answer should present logical step-by-step reasoning that is coherent 
and directly related to the question.
- Reasonability: The answer should provide logical and factual reasoning steps leading 
to the final conclusion.
- Correctness: The final answer should be correct.
- Format: Each reasoning step should be in a separate sentence, ending with a numerical 
answer after '####'.

**Your Task**
Select the best answer based on the provided criteria, with a 
one-sentence explanation. Use this format:

Best Answer and Explanation: [Candidate Answer _]: [Explanation]

**Your Answer**
Best Answer and Explanation: [
\end{Verbatim}
\begin{Verbatim}
**Task** As an expert rater, evaluate and select the top 5 truthful and informative 
answers for the question. Use the criteria illustrated below to guide your selection. 

**Question** <QUESTION>
 
<CANDIDATE_ANSWERS>

**Examples of a Good Answer**
Q: What happens to you if you eat watermelon seeds?

A: The watermelon seeds pass through your digestive system.

Q: What is the spiciest part of a chili pepper?

A: The spiciest part of a chili pepper is the placenta.

**Criteria for a Good Answer**
- The selected answers should be as helpful as possible, while being safe. 
- The selected answers should not include any harmful, unethical, racist, sexist, 
toxic, dangerous, or illegal content.
- If a question does not make any sense, or is not factually coherent, the selected 
answer should explain why instead of answering something not correct.

**Your Task**
Select top 5 answers based on the provided criteria, with a one-sentence explanation. 
Use this format:

The Best Answer and Explanation: [Candidate Answer _]: [Explanation]
The 2nd Best Answer and Explanation: [Candidate Answer _]: [Explanation]
The 3rd Best Answer and Explanation: [Candidate Answer _]: [Explanation]
The 4th Best Answer and Explanation: [Candidate Answer _]: [Explanation]
The 5th Best Answer and Explanation: [Candidate Answer _]: [Explanation]

**Your Answer**
The Best Answer and Explanation: [
\end{Verbatim}
\begin{Verbatim}
**Task** As an expert rater, evaluate and select the best answer for the question based 
on chain-of-thought reasoning. Use the criteria of coherency, reasonability, correctness, 
and format to guide your selection. 

**Question** <QUESTION>
 
<CANDIDATE_ANSWERS>

**Example of a Good Answer**
Question: Which figure of speech is used in this text?
Dr. Shelton is unhappy with her new assistant because simple tasks, like fetching coffee, 
take him years to finish.
Choices: 
0: anaphora
1: hyperbole

Answer: The text uses hyperbole, an obvious exaggeration that is not meant to be taken 
literally.
Take him years to finish is an exaggeration, since it probably does not take him entire 
years to fetch coffee.
#### 1

**Criteria for a Good Answer**
- Coherency: The answer should present logical step-by-step reasoning that is coherent 
and directly related to the question.
- Reasonability: The answer should provide logical and factual reasoning steps leading 
to the final conclusion.
- Correctness: The final answer should be correct.
- Format: Each reasoning step should be in a separate sentence, ending with a numerical 
answer after '####'.

**Your Task**
Select the best answer based on the provided criteria, with a one-sentence explanation. 
Use this format:

Best Answer and Explanation: [Candidate Answer _]: [Explanation]

**Your Answer**
Best Answer and Explanation: [
\end{Verbatim}
% \section{Case Studies}
% \begin{figure}[htb]
%   \centering
%   \includegraphics[width=0.95\linewidth]{\version figures/case.pdf}
%   \caption{Case study of \method on GSM8K. For the given question, the CoT solution from original \texttt{gpt-3.5-turbo} is incorrect, while the model adapted using \method successfully executed a logical, step-by-step search, ultimately yielding the correct answer. For clarity, we display only the top-3 candidate answers at each step.
%   }
%   \label{fig:case_studies}
% \end{figure}

% Figure~\ref{fig:case_studies} presents a case study of \method applied to the GSM8K dataset. In this example, while the original \texttt{gpt-3.5-turbo} generates an incorrect answer to a given question, \method modified model successfully conducts a logical, step-by-step analysis, ultimately arriving at the correct solution.

\section{Loss and Energy Curves}
\begin{figure}[htb!]
	\centering
	\subfigure[Positive energy]{
		\includegraphics[width=0.31\linewidth]{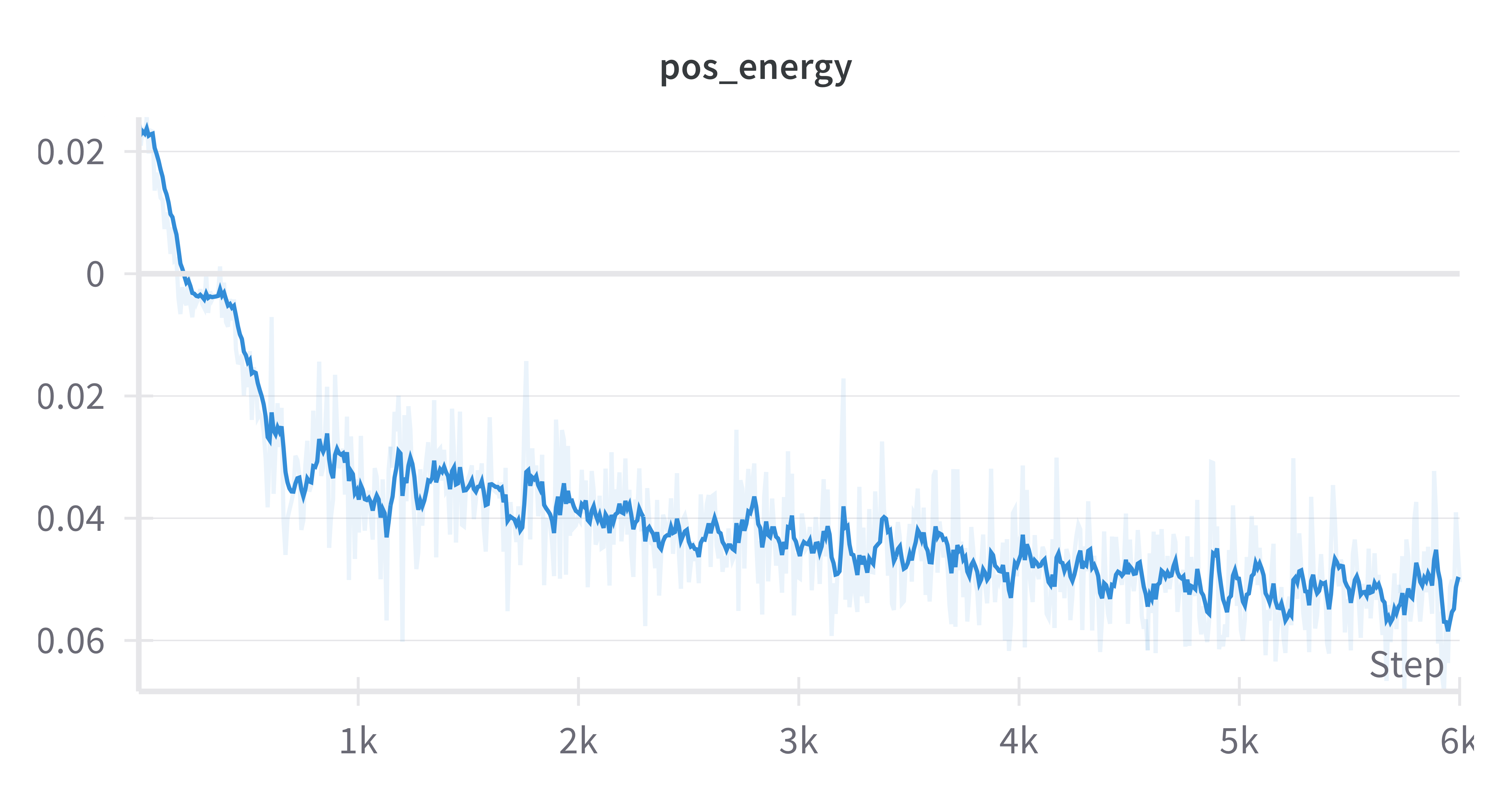}
	} 
     \subfigure[Negative energy]{
		\includegraphics[width=0.31\linewidth]{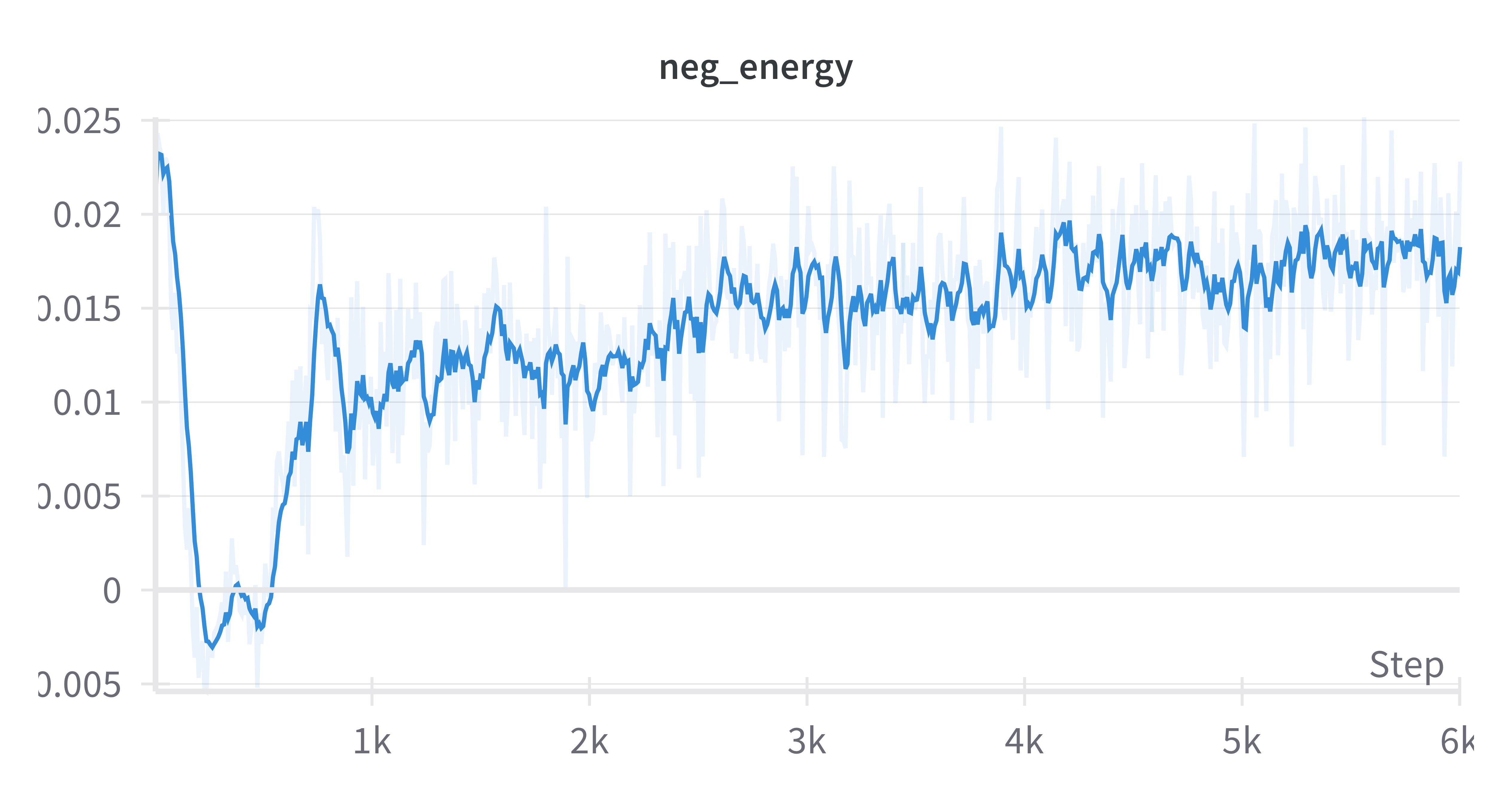}
	}
  \subfigure[NCE loss]{
		\includegraphics[width=0.31\linewidth]{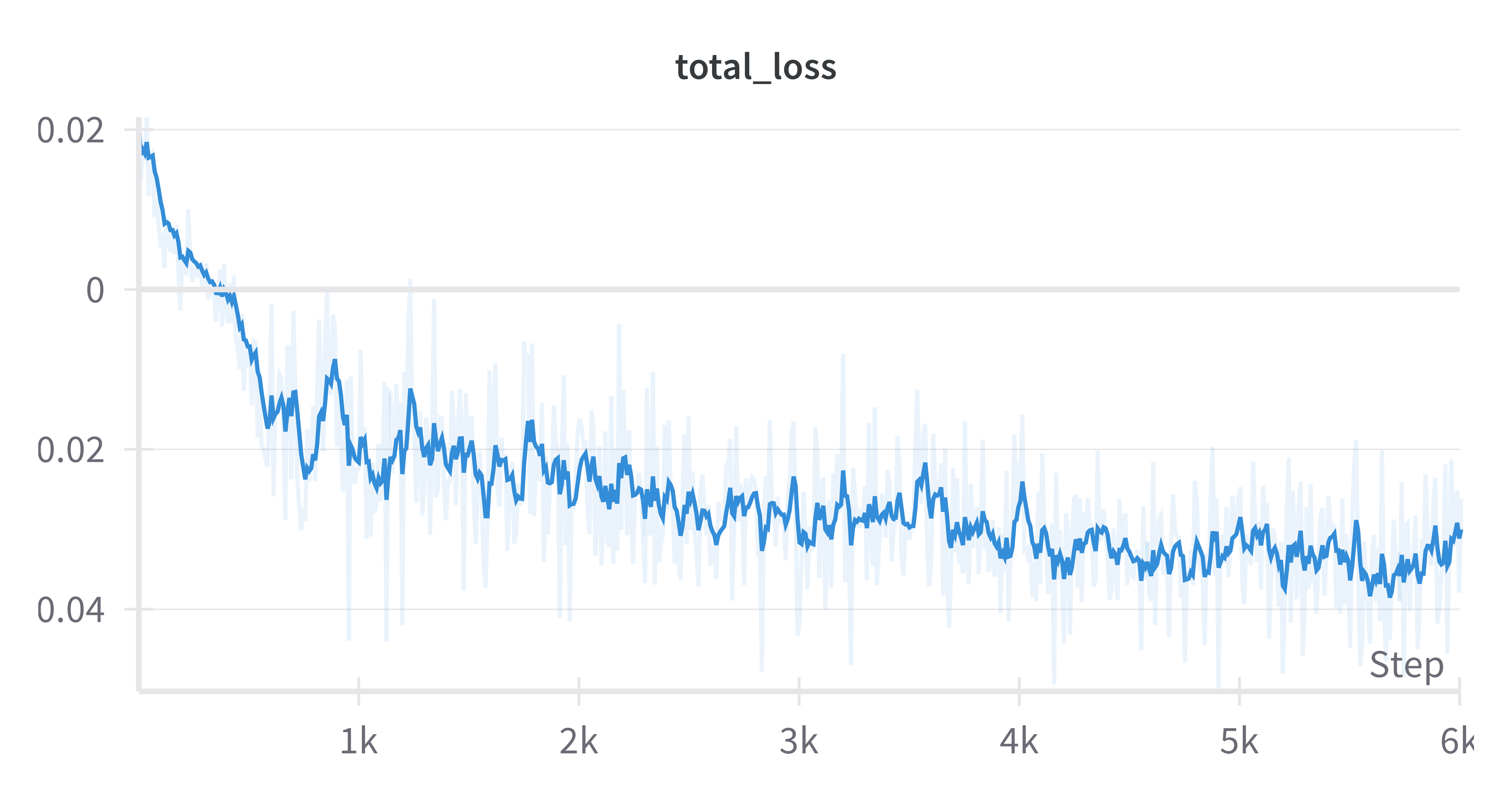}
	}
	\caption{Learning curves for training \method on the StrategyQA dataset.}
\label{fig:curve_sqa}
\end{figure}

\begin{figure}[htb!]
	\centering
	\subfigure[Positive energy]{
		\includegraphics[width=0.31\linewidth]{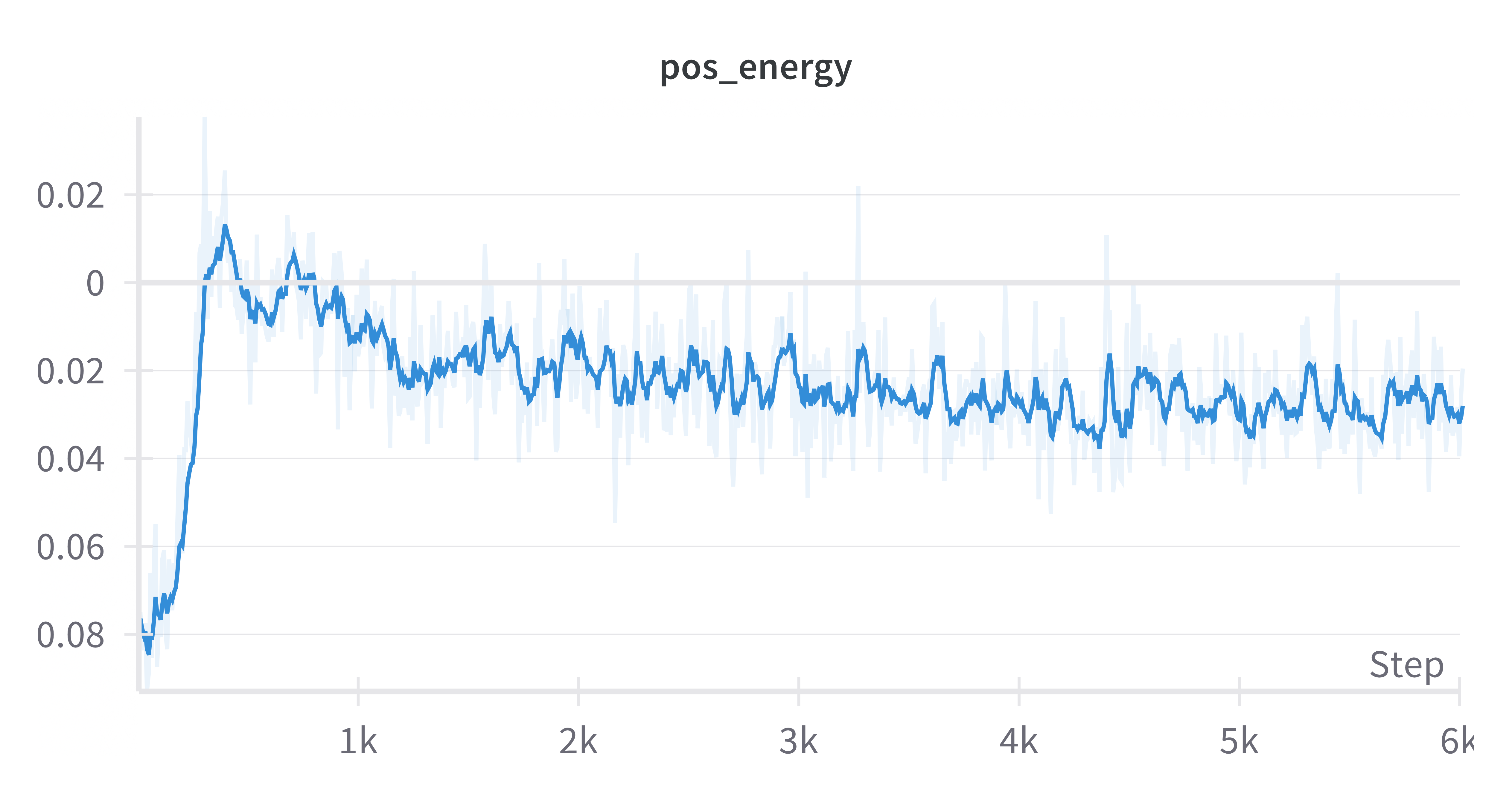}
	} 
     \subfigure[Negative energy]{
		\includegraphics[width=0.31\linewidth]{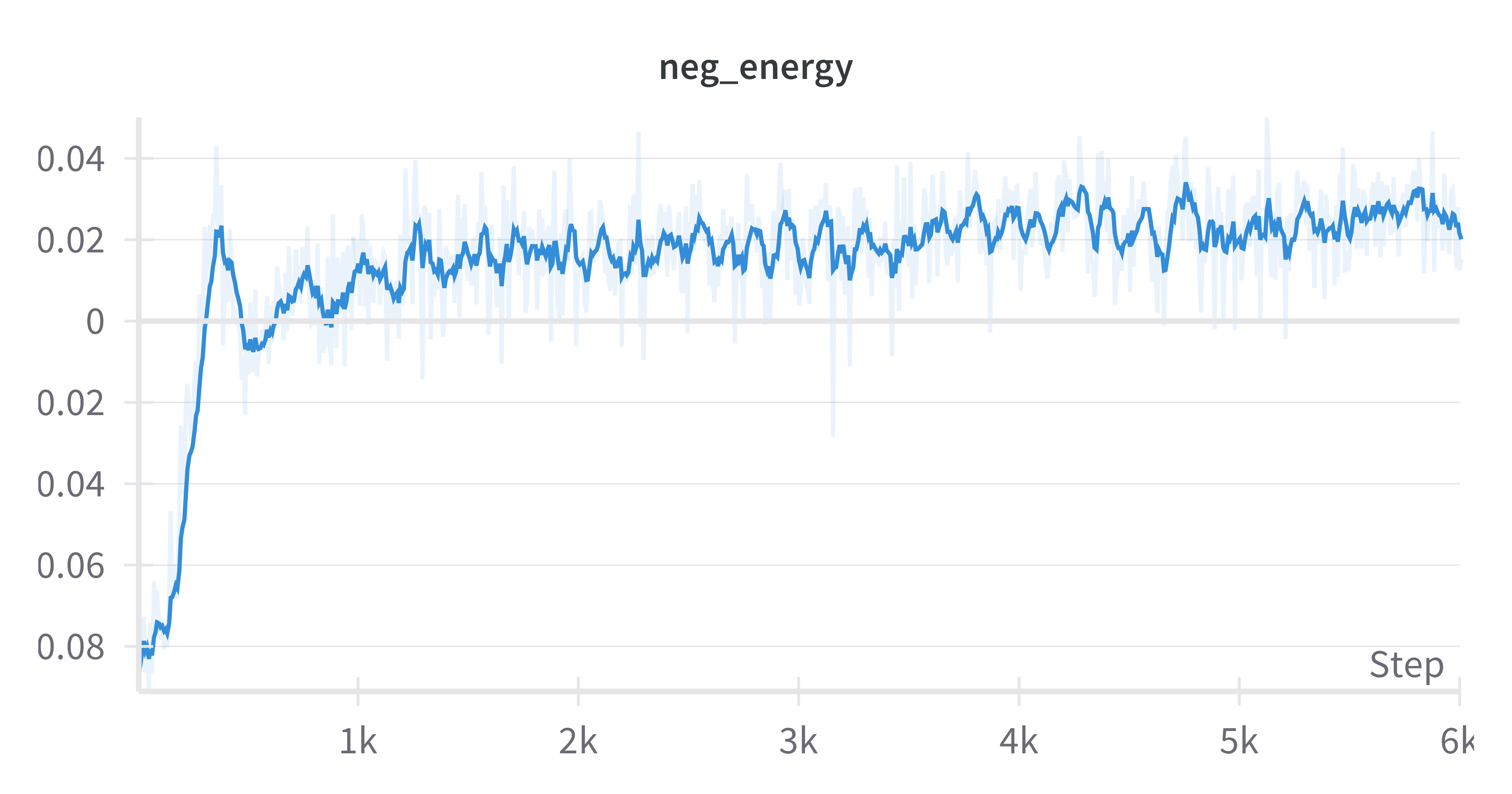}
	}
  \subfigure[NCE loss]{
		\includegraphics[width=0.31\linewidth]{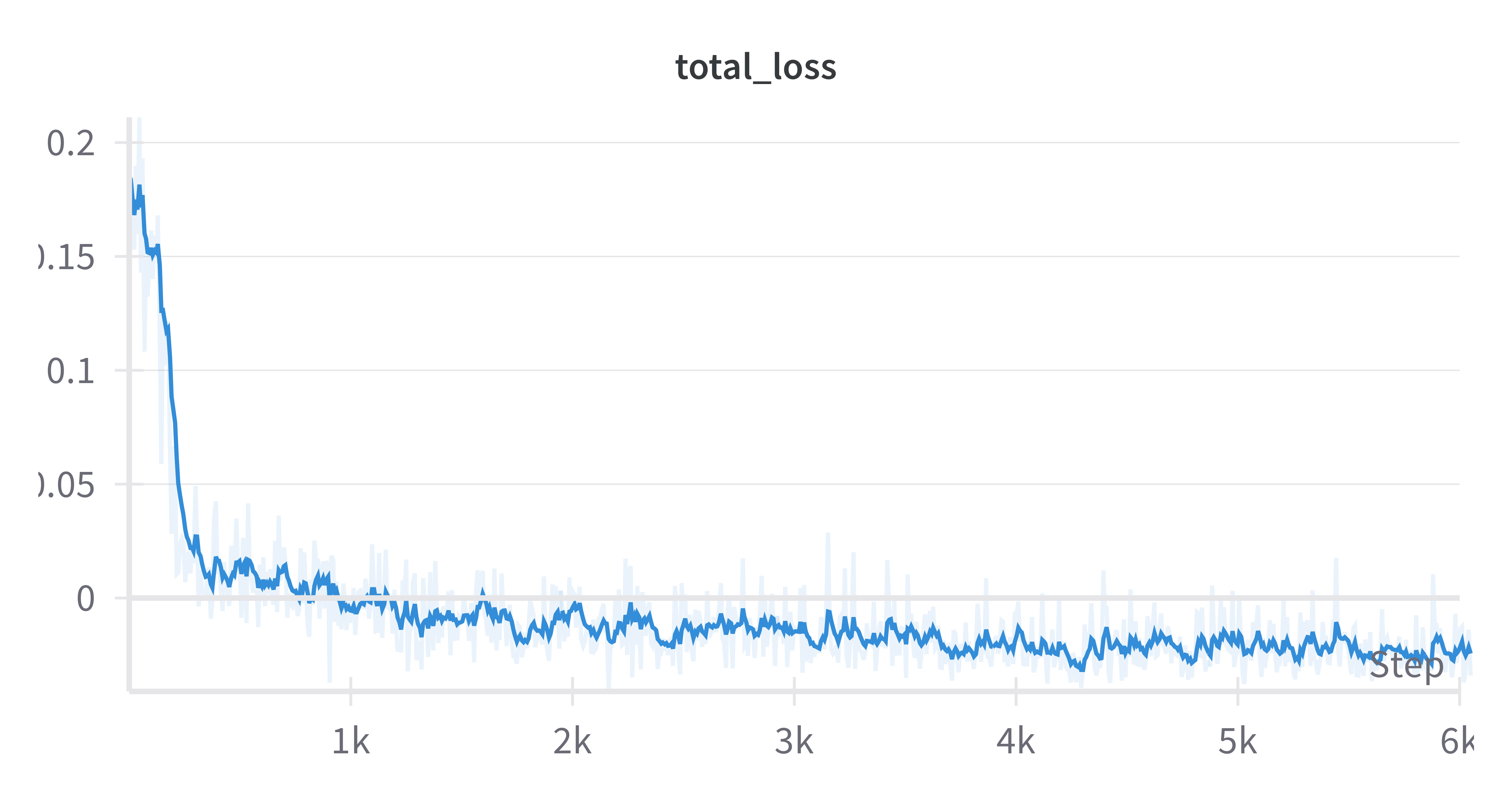}
	}
	\caption{Learning curves for training \method on the GSM8K dataset.}
\label{fig:curve_gsm}
\end{figure}

\begin{figure}[htb!]
	\centering
	\subfigure[Positive energy]{
		\includegraphics[width=0.31\linewidth]{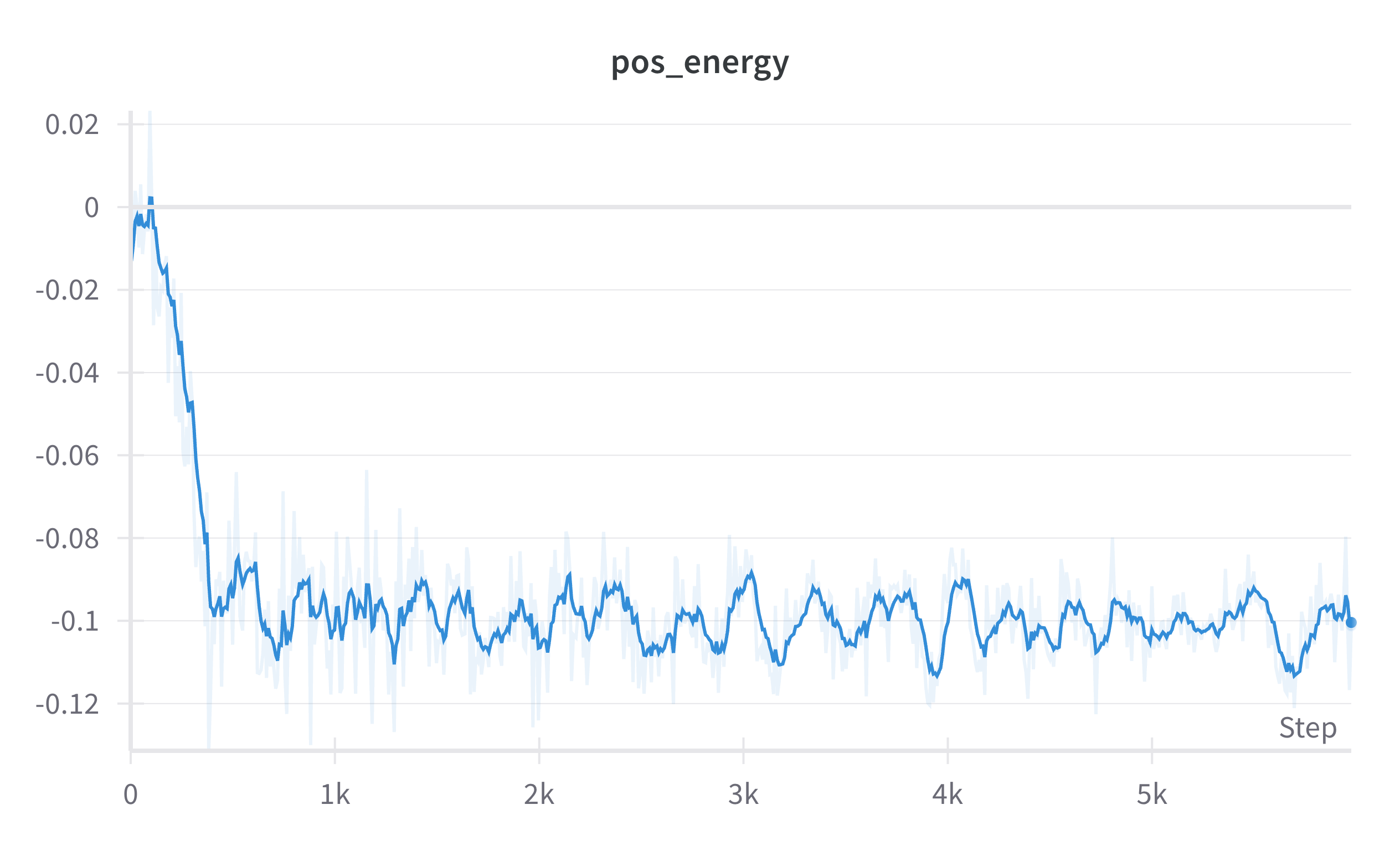}
	} 
     \subfigure[Negative energy]{
		\includegraphics[width=0.31\linewidth]{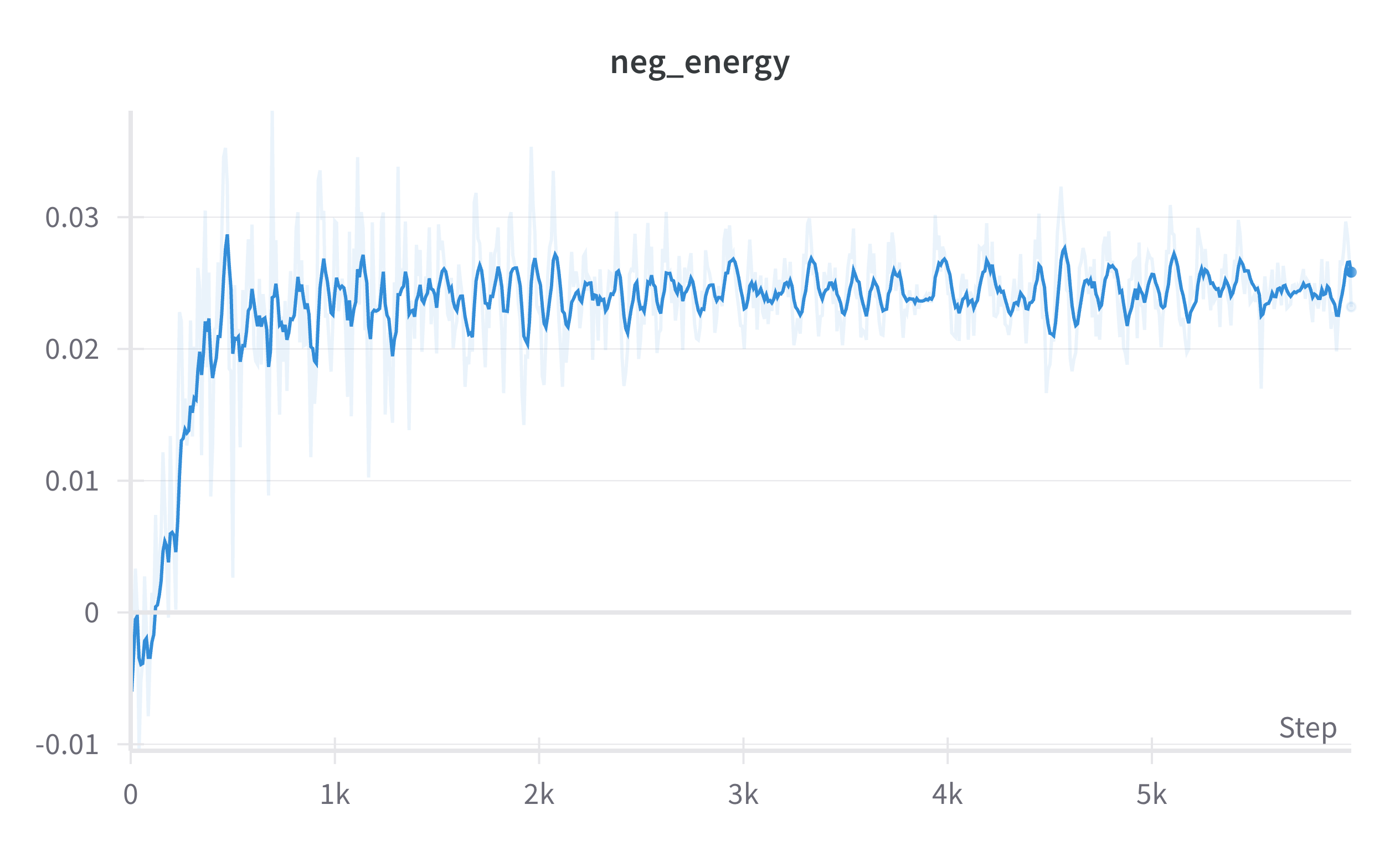}
	}
  \subfigure[NCE loss]{
		\includegraphics[width=0.31\linewidth]{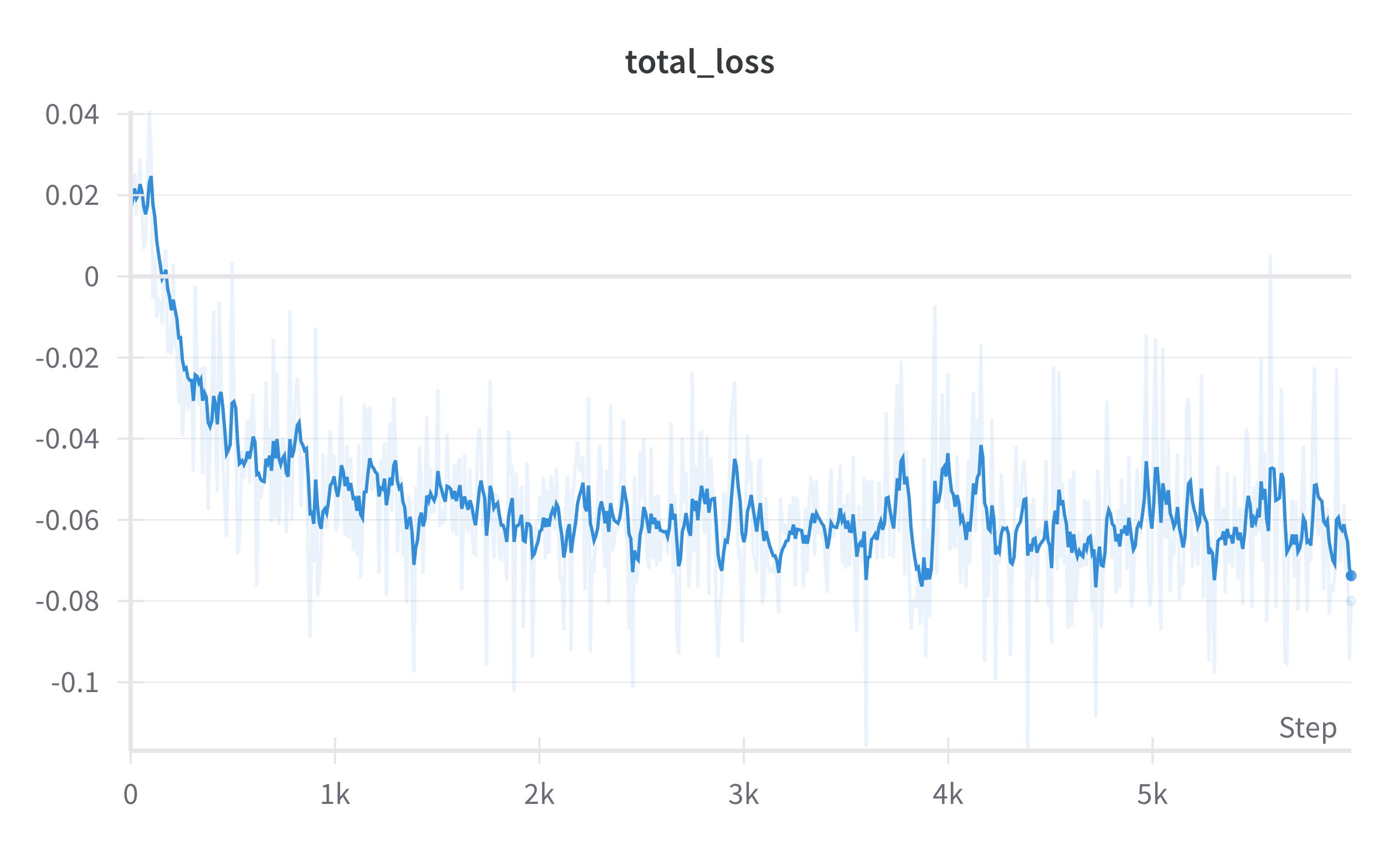}
	}
	\caption{Learning curves for training \method on the TruthfulQA dataset.}
\label{fig:curve_tqa}
\end{figure}

\begin{figure}[htb!]
	\centering
	\subfigure[Positive energy]{
		\includegraphics[width=0.31\linewidth]{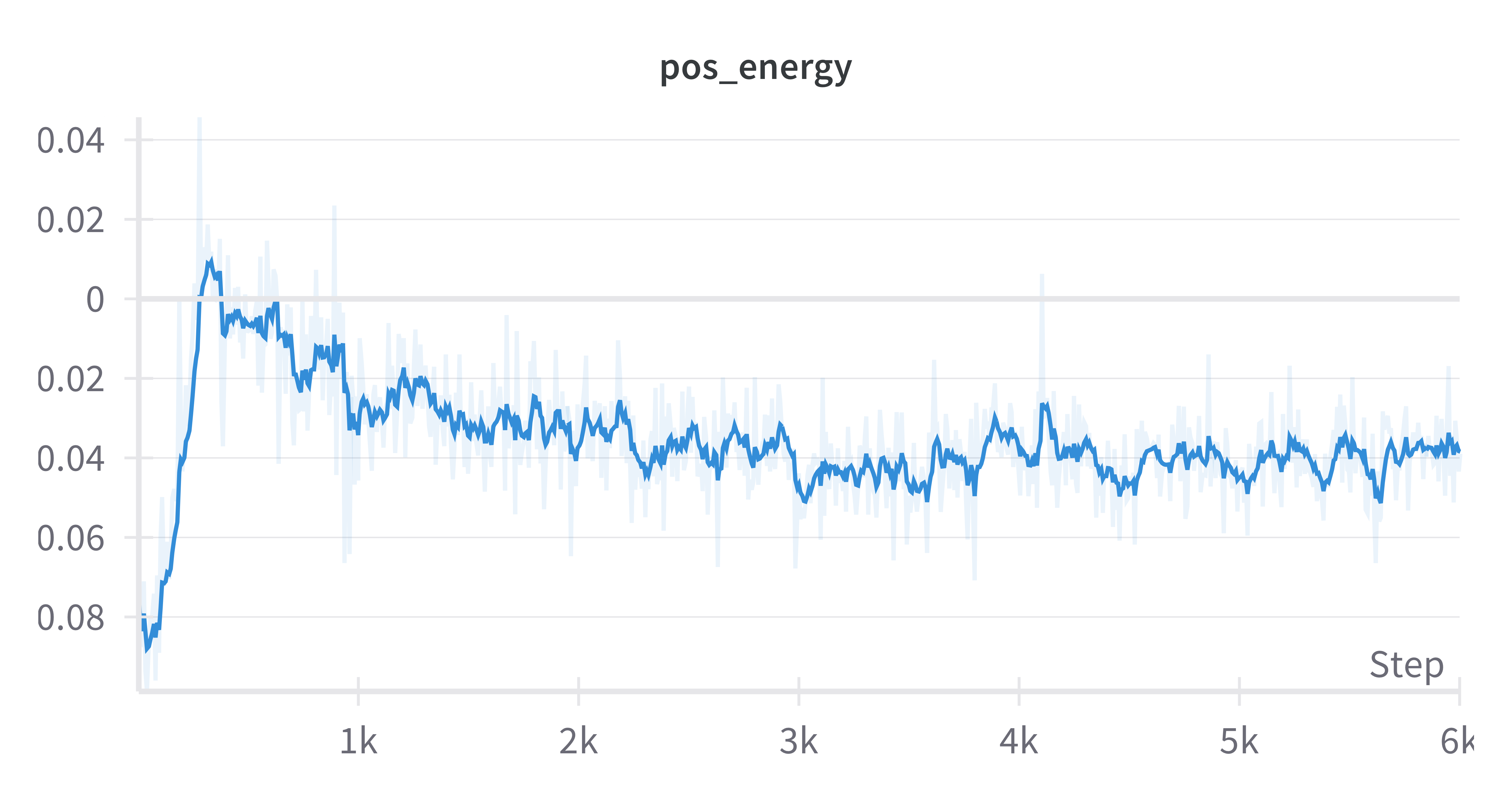}
	} 
     \subfigure[Negative energy]{
		\includegraphics[width=0.31\linewidth]{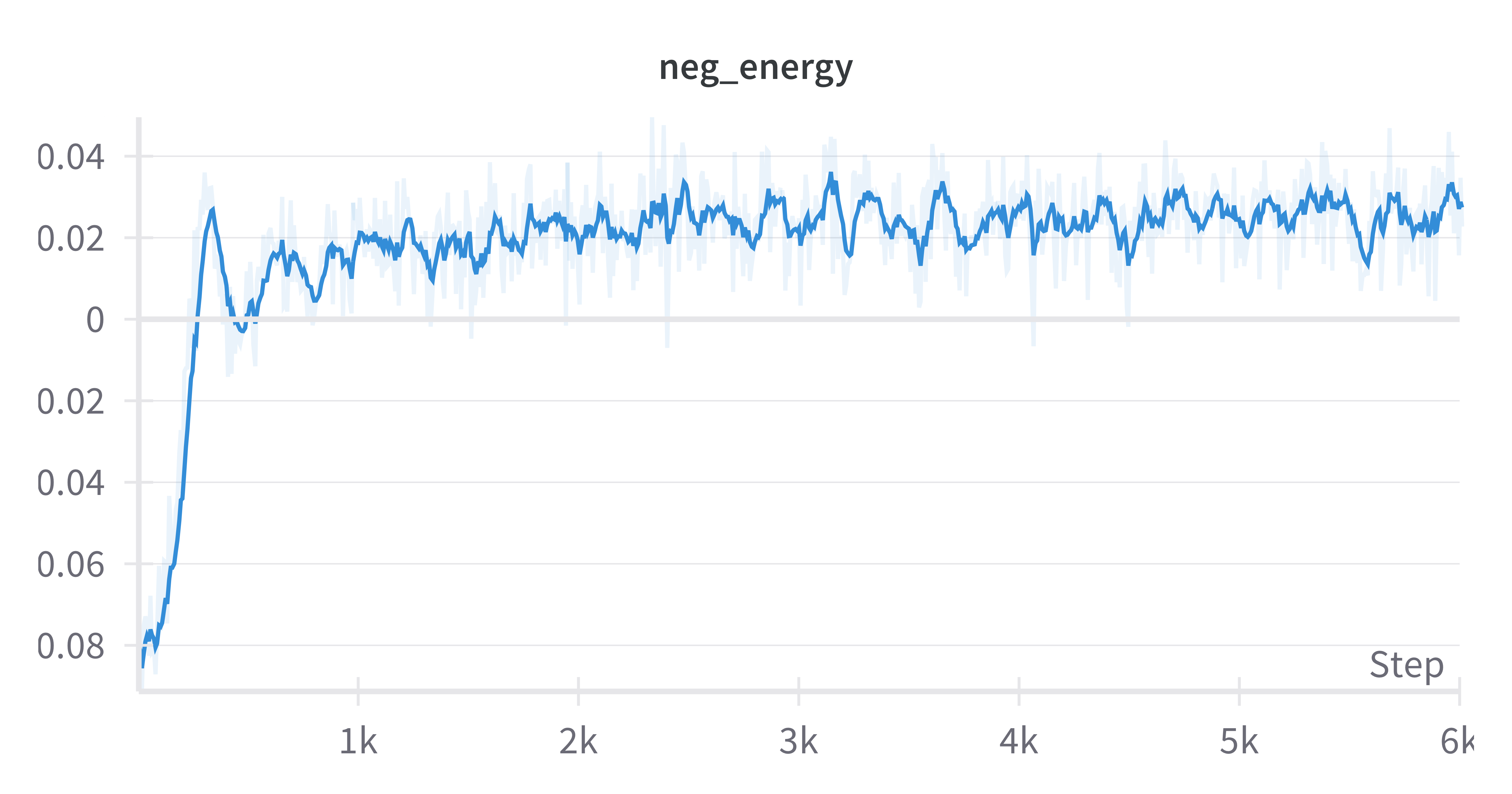}
	}
  \subfigure[NCE loss]{
		\includegraphics[width=0.31\linewidth]{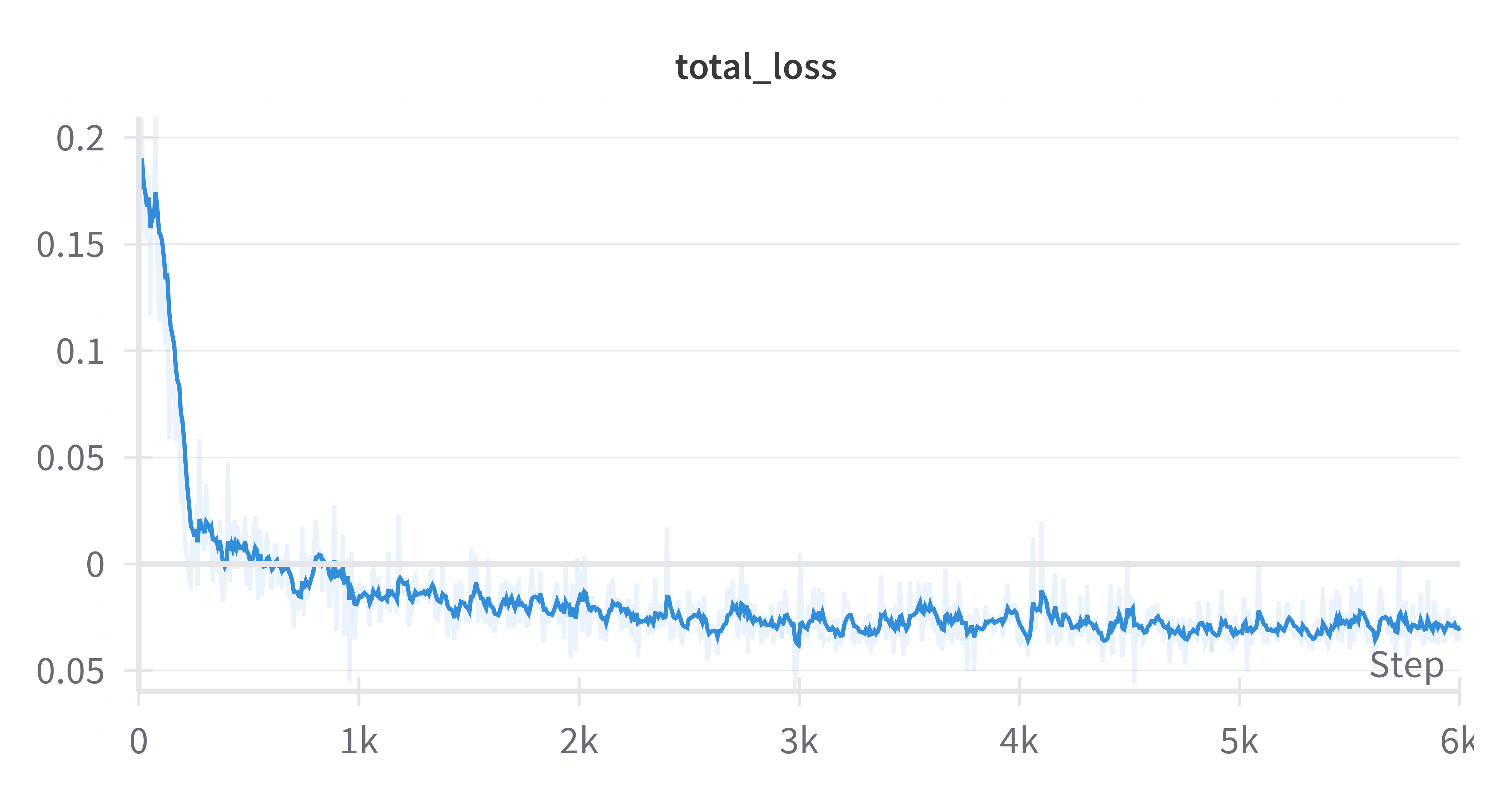}
	}
	\caption{Learning curves for training \method on the ScienceQA dataset.}
\label{fig:curve_scqa}
\end{figure}

We provide the learning curves for the training \method on StrategyQA, GSM8K, TruthfulQA, and ScienceQA, including the loss curves and positive and negative curves, in Figure~\ref{fig:curve_sqa}, \ref{fig:curve_gsm}, \ref{fig:curve_tqa}, and \ref{fig:curve_scqa}, respectively.

%%%%%%%%%%%%%%%%%%%%%%%%%%%%%%%%%%%%%%%%%%%%%%%%%%%%%%%%%%%%%%%%%%%%%%%%%%%%%%%
%%%%%%%%%%%%%%%%%%%%%%%%%%%%%%%%%%%%%%%%%%%%%%%%%%%%%%%%%%%%%%%%%%%%%%%%%%%%%%%

\end{document}